\documentclass[final]{cvpr}

\usepackage{times}
\usepackage{epsfig}
\usepackage{graphicx}
\usepackage{amsmath}
\usepackage{amssymb}
\usepackage{multirow}
\usepackage{pifont}
\usepackage{array}
\usepackage{setspace}
\usepackage{comment}
\usepackage{color}
\newcommand{\tabincell}[2]{\begin{tabular}{@{}#1@{}}#2\end{tabular}}

\usepackage[pagebackref=true,breaklinks=true,letterpaper=true,colorlinks,bookmarks=false]{hyperref}

\def\cvprPaperID{5692} 
\def\confYear{CVPR 2021}

\begin{document}
	
	\title{Unsupervised Degradation Representation Learning for Blind Super-Resolution}
	
\author{Longguang Wang$^{1}$, Yingqian Wang$^{1}$, Xiaoyu Dong$^{2,3}$, Qingyu Xu$^{1}$, Jungang Yang$^{1}$, Wei An$^{1}$, Yulan Guo$^{1*}$\\
	$^{1}$National University of Defense Technology~~~~~
	$^{2}$The University of Tokyo~~~~~~
	$^{3}$RIKEN AIP\\
	{\tt\small \{wanglongguang15,yulan.guo\}@nudt.edu.cn}}
	
	\maketitle
	\pagestyle{empty}
	\thispagestyle{empty}
	
	\begin{abstract}
		Most existing CNN-based super-resolution (SR) methods are developed based on an assumption that the degradation is fixed and known ({e.g.}, bicubic downsampling). However, these methods suffer a severe performance drop when the real degradation is different from their assumption. To handle various unknown degradations in real-world applications, previous methods rely on degradation estimation to reconstruct the SR image.
		Nevertheless, degradation estimation methods are usually time-consuming and may lead to SR failure due to large estimation errors. In this paper, we propose an unsupervised degradation representation learning scheme for blind SR without explicit degradation estimation. Specifically, we learn abstract representations to distinguish various degradations in the representation space rather than explicit estimation in the pixel space. Moreover, we introduce a Degradation-Aware SR (DASR) network with flexible adaption to various degradations based on the learned representations. It is demonstrated that our degradation representation learning scheme can extract discriminative representations to obtain accurate degradation information. Experiments on both synthetic and real images show that our network achieves state-of-the-art performance for the blind SR task. Code is available at: \url{https://github.com/LongguangWang/DASR}.
	\end{abstract}
	
	\section{Introduction}
	Single image super-resolution (SR) aims at recovering a high-resolution (HR) image from a low-resolution (LR) observation. 
	Recently, CNN-based methods \cite{Dong2014Learning,Kim2016Deeply,Lim2017Enhanced,Ahn2018Fast,Qiu2019Embedded} have dominated the research of SR due to the powerful feature representation capability of deep neural networks. 
	As a typical inverse problem, SR is highly coupled with the degradation model \cite{Bell-Kligler2019Blind}. Most existing CNN-based methods are developed based on an assumption that the degradation is known and fixed (\emph{e.g.}, bicubic downsampling). However, these methods suffer a severe performance drop when the real degradation differs from their assumption \cite{Gu2019Blind}. 
	
	To handle various degradations in real-world applications, several methods \cite{Zhang2017Learning,Shocher2018Zero,Zhang2020Deep,Soh2020Meta} have been proposed to investigate the non-blind SR problem. Specifically, these methods use a set of degradations (\emph{e.g.}, different combinations of Gaussian blurs, motion blurs and noises) for training and assume the degradation of the test LR image is known at the inference time. 
	These non-blind methods produce promising SR results when the true degradation is known in priori. 
	
	To super-resolve real images with unknown degradations, degradation estimation \cite{Michaeli2013Nonparametric,Bell-Kligler2019Blind} needs to be performed to provide degradation information for non-blind SR networks \cite{Zhang2017Learning,Shocher2018Zero,Zhang2020Deep,Soh2020Meta}. However, these non-blind methods are  sensitive to degradation estimation. Consequently, the estimation error can further be magnified by the SR network, resulting in obvious artifacts \cite{Gu2019Blind}. To address this problem, Gu \emph{et al.} \cite{Gu2019Blind} proposed an iterative kernel correction (IKC) method to correct the estimated degradation by observing previous SR results. By iteratively correcting the degradation, artifact-free results can be gradually produced. 
	Since numerous iterations are required at test time by degradation estimation methods \cite{Michaeli2013Nonparametric,Bell-Kligler2019Blind} and IKC \cite{Gu2019Blind}, these methods are time-consuming. 
	
	\begin{figure}
		\centering
		\includegraphics[width=1\linewidth]{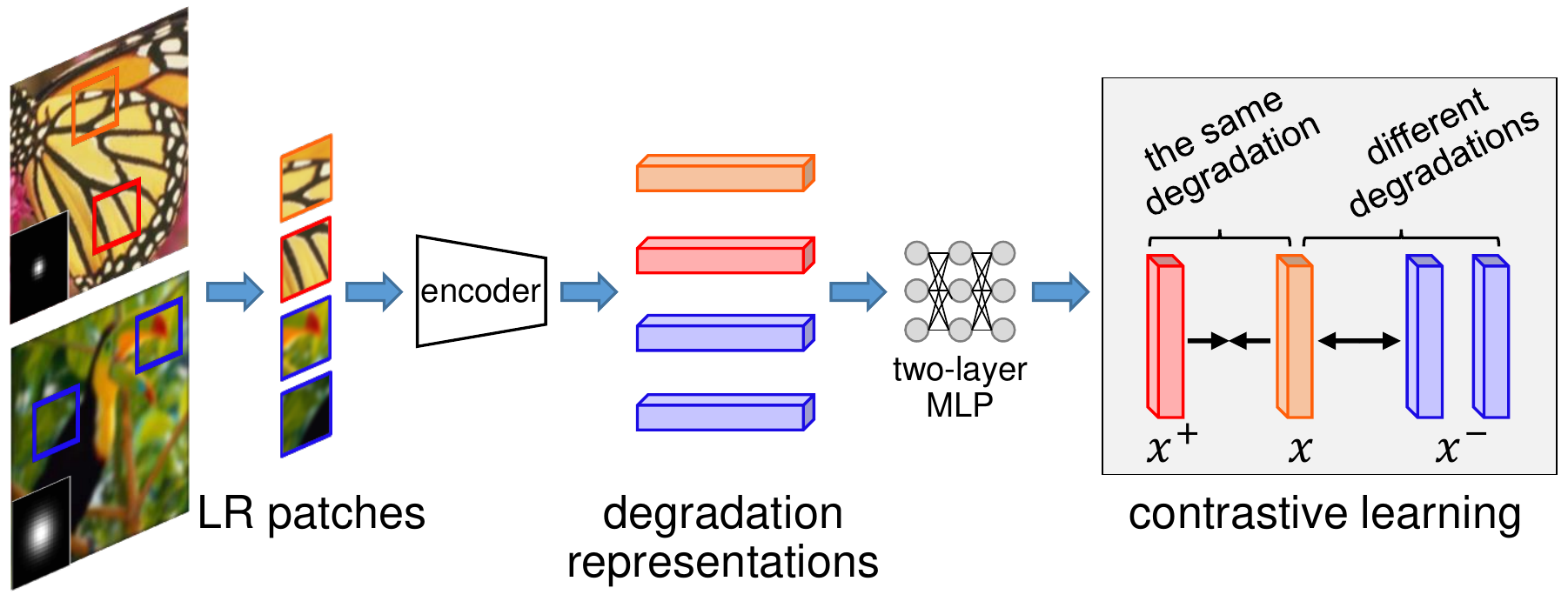}
		\caption{An illustration of our unsupervised degradation representation learning scheme.}
		\label{fig1}
		\vspace{-0.35cm}
	\end{figure}

	Unlike the above methods that explicitly estimate the degradation from an LR image, we investigate a different approach by learning a degradation representation  to distinguish the latent degradation from other ones. 
	Motivated by recent advances of contrastive learning \cite{Hadsell2006Dimensionality,Dosovitskiy2014Discriminative,Tian2019Contrastive,He2020Momentum,Chen2020simple}, a contrastive loss is used to conduct unsupervised degradation representation learning by contrasting positive pairs against negative pairs in the latent space (Fig.~\ref{fig1}).
	The benefits of degradation representation learning are twofold: 
	\textbf{First}, 
	compared to extracting full representations to estimate degradations, it is easier to learn abstract  representations to distinguish different degradations.
	Consequently, we can obtain a discriminative degradation representation to provide accurate degradation information at a single inference. 
	\textbf{Second}, degradation representation learning does not require the supervision from groundtruth degradation. Thus, it can be conducted in an unsupervised manner and is more suitable for real-world applications with unknown degradations.

	In this paper, we introduce an unsupervised degradation representation learning scheme for blind SR. Specifically, we assume the degradation is the same in an image but can vary for different images, which is the general case widely used in  literature \cite{Zhang2017Learning,Bell-Kligler2019Blind,Zhang2020Deep}. Consequently, an image patch should be similar to other patches in the same image (\emph{i.e.}, with the same degradation) and dissimilar to patches from other images (\emph{i.e.}, with different degradations) in the degradation representation space, as illustrated in Fig.~\ref{fig1}. Moreover, we propose a degradation-aware SR (DASR) network with flexible adaption to different degradations based on the learned representations. 
	Specifically, our DASR incorporates degradation information to perform feature adaption by predicting convolutional kernels and channel-wise modulation coefficients from the degradation representation.  
	Experimental results show that our network can handle various degradations and produce promising results on both synthetic and real-world images under blind settings. 
	
	
	\section{Related Work}
	In this section, we briefly review several major works for CNN-based single image SR and recent advances of contrastive learning. 
	
	\subsection{Single Image Super-Resolution}
	
	\noindent \textbf{SR with Single Degradation.}
	As a pioneer work, a three-layer network is used in SRCNN \cite{Dong2014Learning} to learn the LR-HR mapping for single image SR. Since then, CNN-based methods have dominated the research of SR due to their promising performance. Kim \emph{et al.} \cite{Kim2016Accurate} proposed a 20-layer network with a residual learning strategy. Lim \emph{et al.} \cite{Lim2017Enhanced} followed the idea of residual learning and modified residual blocks to build a very deep and wide network, namely EDSR. Zhang \emph{et al.} \cite{Zhang2018Residual} then combined residual learning and dense connection to construct a residual dense network (RDN) with over 100 layers. Haris \emph{et al.} \cite{Haris2018Deep} introduced multiple up-sampling/down-sampling layers to provide an error feedback mechanism and used self-corrected features to produce superior results. Recently, channel attention and second-order channel attention are further introduced by RCAN \cite{Zhang2018Image} and SAN \cite{Dai2019Second} to exploit feature correlation for improved performance. 
	
	\noindent \textbf{SR with Multiple Degradations.}
	Although important advances have been achieved by the above SR methods, they are tailored to a fixed bicubic degradation and suffer severe performance drop when the real degradation differs from the bicubic one \cite{Gu2019Blind}. To handle various degradations, several efforts \cite{Zhang2017Learning,Xu2020Unified,Zhang2020Deep,Hussein2020Correction} have been made to investigate the non-blind SR problem. Specifically, degradation is first used as an additional input in SRMD \cite{Zhang2017Learning} to super-resolve LR images under different degradations. Later, dynamic convolutions are further incorporated in UDVD \cite{Xu2020Unified} to achieve better performance than SRMD. Recently, Zhang \emph{et al.} \cite{Zhang2020Deep} developed an unfolding SR network (USRnet) to handle different degradations by alternately solving a data sub-problem and a prior sub-problem. Hussein \emph{et al.} \cite{Hussein2020Correction} introduced a closed-form correction filter to transform an LR image to match the one generated by bicubic degradation. Then, existing networks trained for bicubic degradation can be used to super-resolve the transformed LR image.
	
	Zero-shot methods have also been investigated to achieve SR with multiple degradations. In ZSSR \cite{Shocher2018Zero},  training is conducted at test time using a degradation and an LR image as its input. Consequently, the network can be adapted to the given degradation. However, ZSSR requires thousands of iterations to converge and is quite time-consuming. To address this limitation, optimization-based meta-learning is used in MZSR \cite{Soh2020Meta}  to make the network adaptive to a specific degradation within a few iterations during inference.
	
	Since degradation is used as an input for these aforementioned methods, they highly rely on degradation estimation methods \cite{Michaeli2013Nonparametric,Bell-Kligler2019Blind} for blind SR. Therefore, degradation estimation errors can ultimately introduce undesired artifacts to the SR results \cite{Gu2019Blind}. 
	To address this problem, Gu \emph{et al.} \cite{Gu2019Blind} proposed an iterative kernel correction (IKC) method to correct the estimated degradation by observing previous SR results. Luo \emph{et al.} \cite{Luo2020Unfolding} developed a deep alternating network (DAN) by iteratively estimating the degradation and restoring an SR image. 
	
	\subsection{Contrastive Learning}
	Contrastive learning has demonstrated its effectiveness in unsupervised representation learning. Previous methods \cite{Doersch2015Unsupervised,Zhang2016Colorful,Noroozi2017Representation,Gidaris2018Unsupervised} usually conduct representation learning by minimizing the difference between the output and a fixed target (\emph{e.g.}, the input itself for auto-encoders). Instead of using a pre-defined and fixed target, contrastive learning maximizes the mutual information in a representation space. Specifically, the representation of a query sample should attract positive counterparts while repelling negative counterparts. The positive counterparts can be transformed versions of the input \cite{Wu2018Unsupervised,Chen2020simple,He2020Momentum}, multiple views of the input \cite{Tian2019Contrastive} and neighboring patches in the same image \cite{Oord2018Representation,Henaff2019Data}. In this paper, image patches generated with the same degradation are considered as positive counterparts and contrastive learning is conducted to obtain content-invariant degradation representations, as shown in Fig.~\ref{fig1}.
	
	
	\section{Methodology}
	\subsection{Problem Formulation}
	The degradation model of an LR image $I^{LR}$ can be formulated as follows:
	\begin{equation}
	\label{eq1}
	I^{LR}=(I^{HR}\otimes{k})\downarrow_{s}+n,
	\end{equation}
	where $I^{HR}$ is the HR image, $k$ is a blur kernel, $\otimes$ denotes convolution operation, $\downarrow_{s}$ represents downsampling operation with scale factor $s$ and $n$ usually refers to additive white Gaussian noise. Following \cite{Zhang2017Learning,Gu2019Blind}, we use bicubic downsampler as the downsampling operation. In this paper, we first investigate a noise-free degradation model with isotropic Gaussian kernels and then a more general degradation model with anisotropic Gaussian kernels and noises. Finally, we test our network on real-world degradations.
	
	\subsection{Our Method}
	\begin{figure*}[ht]
		\centering
		\includegraphics[width=1\linewidth]{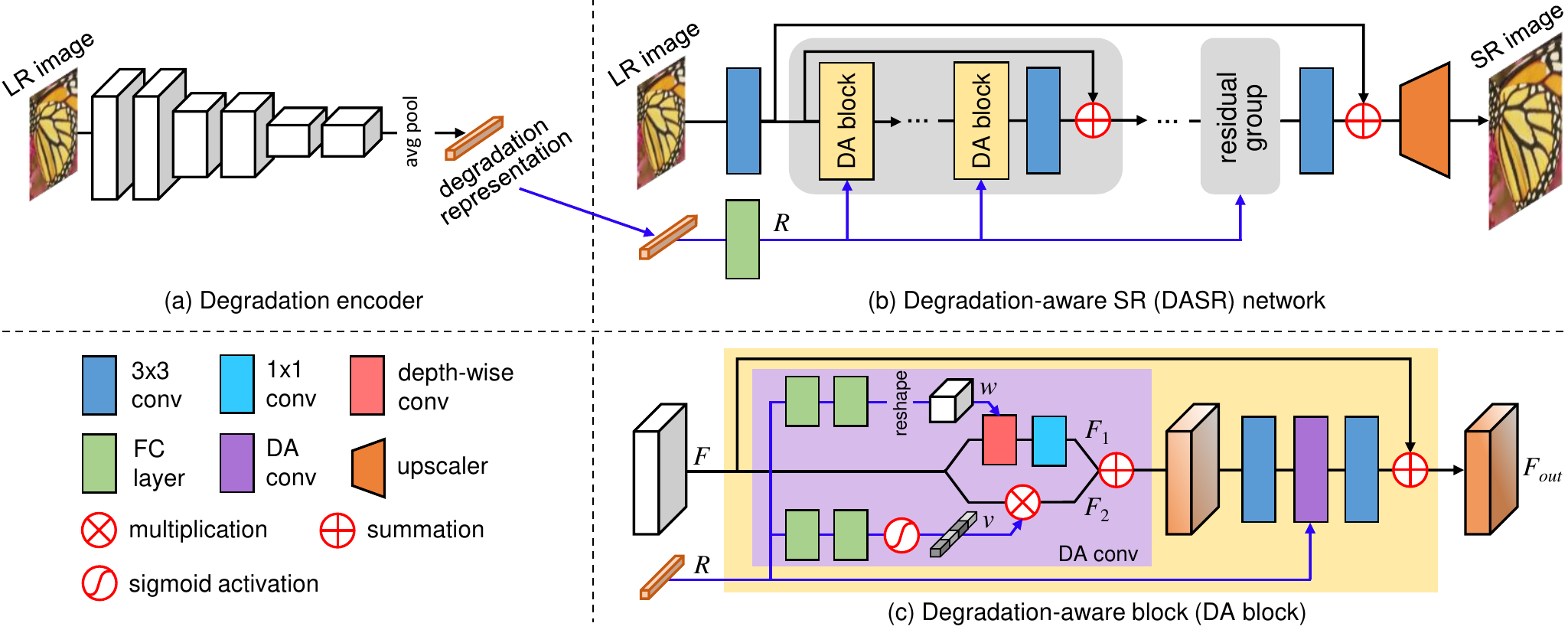}
		\caption{An overview of our blind SR framework.}
		\label{fig2}
		\vspace{-0.35cm}
	\end{figure*}

	Our blind SR framework consists of a degradation encoder and a degradation-aware SR network, as illustrated in Fig.~\ref{fig2}. First, the LR image is fed to the degradation encoder (Fig.~\ref{fig2}(a)) to obtain a degradation representation. Then, this representation is incorporated in the degradation-aware SR network (Fig.~\ref{fig2}(b)) to produce the SR result.
	
	\subsubsection{Degradation Representation Learning}
	\label{sec.3.1.2}
	
	The goal of degradation representation learning is to extract a discriminative representation from the LR image in an unsupervised manner. As shown in Fig.~\ref{fig1}, we use a contrastive learning framework \cite{He2020Momentum} for degradation representation learning. Note that, we assume the degradation is the same in each image and varies for different images.
	
	\noindent \textbf{Formulation.}
	Given an image patch (annotated with an orange box in Fig.~\ref{fig1}) as the query patch, other patches extracted from the same LR image (\emph{e.g.}, the patch annotated with a red box) can be considered as positive samples. In contrast, patches from other LR images (\emph{e.g.}, patches annotated with blue boxes) can be referred to as negative samples. Then, we encode the query, positive and negative patches into degradation representations using a convolutional network with six layers (Fig.~\ref{fig2}(a)). As suggested in SimCLR \cite{Chen2020simple} and MoCo v2 \cite{Chen2020Improved}, the resulting representations are further fed to a two-layer multi-layer perceptron (MLP) projection head to obtain ${{x}},{{x}}^{+}$ and ${{x}}^{-}$. ${{x}}$ is encouraged to be similar to ${{x}}^{+}$ while being dissimilar to ${{x}}^{-}$.   Following MoCo \cite{He2020Momentum}, an InfoNCE loss is used to measure the similarity. That is, 
	\begin{equation}
	L_x=-{\rm log}\frac{{\rm exp}({{x}}\cdot{{{{x}}}^+}/\tau)}{\sum_{n=1}^N{{\rm exp}({{x}}\cdot{{{x}}_n^-}/\tau)}},
	\end{equation}
	where $N$ is the number of negative samples, $\tau$ is a temperature hyper-parameter and $\cdot$ represents the dot product between two vectors.

	As emphasized in existing contrastive learning methods \cite{Chen2020simple,He2020Momentum,Park2020Contrastive}, a large dictionary covering a rich set of negative samples is critical for good representation learning. To obtain content-invariant degradation representations, a queue containing samples with various contents and degradations is maintained. During training, $B$ LR images (\emph{i.e.}, $B$ different degradations) are first randomly selected and then two patches are randomly cropped from each image. Next, these $2B$ patches are encoded into $\{{{p}}_i^{1},{{p}}_i^2\!\in\!{\mathbb{R}^{256}}\}$ using our degradation encoder, where ${{p}}_i^{1}$ is the embedding of the first patch from the $i^{\rm th}$ image. For the $i^{\rm th}$ image, we refer to ${{p}}_i^{1}$ and ${{p}}_i^{2}$ as query and positive samples.
	The overall loss is defined as:
	\begin{equation}
	\label{eq3}
	L_{degrad}=\sum_{i=1}^B-{\rm log}\frac{{\rm exp}({{p}}_i^1\cdot{{{p}}_i^2}/\tau)}{\sum^{N_{queue}}_{j=1}{{\rm exp}({{p}}_i^1\cdot{{{p}}^j_{queue}}/\tau)}},
	\end{equation}
	where $N_{queue}$ is the number of samples in the queue and ${{p}}^j_{queue}$ represents the $j^{\rm th}$ negative sample.
	
	\noindent \textbf{Discussion.}
	Existing degradation estimation methods \cite{Michaeli2013Nonparametric,Bell-Kligler2019Blind,Gu2019Blind} aim at estimating the degradation (usually the blur kernel) at pixel level.
	That is, these methods learn to extract full representations of the degradation. However, they are time-consuming as numerous iterations are required during inference. For example, KernelGAN conducts network training during test and takes over 60 seconds for a single image \cite{Bell-Kligler2019Blind}. 
	Different from these methods, we aim at learning a ``good'' abstract representation to distinguish a specific degradation from others rather than explicitly estimating the degradation. It is demonstrated in Sec.~\ref{Sec4.2} that our degradation representation learning scheme is effective yet efficient and can obtain discriminative representations at a single inference. Moreover, our scheme does not require the supervision from groundtruth degradation and can be conducted in an unsupervised manner.

	\subsubsection{Degradation-Aware SR Network}
	
	With degradation representation learning, a degradation-aware SR (DASR) network is proposed to super-resolve the LR image using the resultant representation, as shown in Fig.~\ref{fig2}(b).
	
	\noindent \textbf{Network Architecture.}
	Figure~\ref{fig2}(b) illustrates the architecture of our DASR network. Degradation-aware block (DA block) is used as the building block and the high-level structure of RCAN \cite{Zhang2018Image} is employed. Our DASR network consists of 5 residual groups, with each group comprising of 5 DA blocks. 
	
	Within each DA block, two DA convolutional layers are used to adapt the features based on the degradation representation, as shown in Fig.~\ref{fig2}(c). 
	Motivated by the observation that convolutional kernels of models trained for different restoration levels share similar patterns but have different statistics \cite{He2019Modulating}, our DA convolutional layer learns to predict the kernel of a depth-wise convolution conditioned on the degradation representation. Specifically, the degradation representation $R$ is fed to two full-connected (FC) layers and a reshape layer to produce a convolutional kernel $w\in{\mathbb{R}^{C\times1\times3\times3}}$. Then, the input feature $F$ is processed with a $3\times3$ depth-wise convolution (using $w$) and a $1\times1$ convolution to produce $F_1$. 
	Moreover, inspired by CResMD \cite{He2020Interactive} (that uses controlling variables to rescale different channels to handle multiple degradations), our DA convolutional layer also learns to generate modulation coefficients based on the degradation representation to perform channel-wise feature adaption. Specifically, $R$ is passed to another two FC layers and a sigmoid activation layer to generate channel-wise modulation coefficients $v$. Then, $v$ is used to rescale different channel components in $F$, resulting in $F_2$. Finally, $F_1$ is summed up with $F_2$ and fed to the subsequent layers to produce the output feature $F_{out}$. 
	
	\noindent \textbf{Discussion.}
	Existing SR networks \cite{Zhang2017Learning,Xu2020Unified} for multiple degradations commonly concatenate degradation representations with image features and feed them to CNNs to exploit degradation information. However, due to the domain gap between degradation representations and image features, directly processing them as a whole using convolution will introduce interference \cite{Gu2019Blind}. Different from these networks, by learning to predict convolutional kernels and modulation coefficients based on the degradation representations, our DASR can well exploit degradation information to adapt to specific degradations. 
	It is demonstrated in Sec.~\ref{Sec4.2} that our DASR benefits from DA convolution to achieve flexible adaption to various degradations with better SR performance.

	\section{Experiments}
	\subsection{Datasets and Implementation Details}
	We synthesized LR images according to Eq.~\ref{eq1} for training and test. Following \cite{Gu2019Blind}, we used 800 training images in DIV2K \cite{Agustsson2017NTIRE} and 2650 training images in Flickr2K \cite{Timofte2017NTIRE} as the training set, and included four benchmark datasets (Set5 \cite{Bevilacqua2012Low}, Set14 \cite{Zeyde2010Single}, B100 \cite{Martin2001database} and Urban100 \cite{Huang2015Single}) for evaluation. The size of the Gaussian kernel was fixed to $21\times21$ following \cite{Gu2019Blind}. We first trained our network on noise-free degradations with isotropic Gaussian kernels only. The ranges of kernel width $\sigma$ were set to [0.2,2.0], [0.2,3.0] and [0.2,4.0] for $\times2/3/4$ SR, respectively. 
	Then, our network was trained on more general degradations with anisotropic Gaussian kernels and noises. Anisotropic Gaussian kernels characterized by a Gaussian probability density function $N(0,\Sigma)$ (with zero mean and varying covariance matrix $\Sigma$) were considered. The covariance matrix $\Sigma$ was determined by two random eigenvalues $\lambda_1,\lambda_2\sim{U(0.2,4)}$ and a random rotation angle $\theta\sim{U(0,\pi)}$. The range of noise level was set to $[0, 25]$.

	\begin{table*}[t]
		\caption{PSNR results achieved on Set14 for $\times4$ SR. *: SRMDNF is a version of SRMD trained with noise-free samples.}
		\label{tab0}
		\begin{center}
			\footnotesize
			\setlength{\tabcolsep}{1.2mm}{
				\begin{tabular}{|l|c|cc|c|c|ccccc|}
					\hline
					\multirow{2}{*}{Method}
					& \multirow{2}{*}{\tabincell{c}{Degradation\\Representation Learning}}
					& \multicolumn{2}{c|}{{DA Conv}}
					& \multirow{2}{*}{\tabincell{c}{Oracle\\Degradation}}
					& \multirow{2}{*}{Time}
					& \multicolumn{5}{c|}{Kernel Width $\sigma$}
					\tabularnewline
					&& \tabincell{c}{Kernel} & \tabincell{c}{Modulation}  &&& 0.2 & 1.0 & 1.8 & 2.6 & 3.4
					\tabularnewline
					\hline
					SRMDNF* \cite{Zhang2017Learning}   & - & - & - & \ding{51}
					& 3ms & 28.44 & 28.50 & 28.49 & 28.31 & 27.55
					\tabularnewline
					SRMDNF \cite{Zhang2017Learning} + KernelGAN \cite{Bell-Kligler2019Blind} & - & - & - & \ding{55}
					& 3ms+190s & 26.62 & 26.74 & 26.62 & 26.88 & 26.66
					\tabularnewline
					SRMDNF \cite{Zhang2017Learning} + Predictor \cite{Gu2019Blind} & - & - & - & \ding{55}
					& 3ms+2ms & 26.13 & 26.15 & 26.19 & 26.20 & 26.18
					\tabularnewline
					\hline
					Model 1 & \ding{55} & \ding{51} & \ding{51} & \ding{55}
					& 70ms & 28.46 & 28.40 & 28.30 & 27.77 & 26.79
					\tabularnewline
					Model 2  & \ding{51} & \ding{55} & \ding{55} & \ding{55}
					& 51ms & 28.49 & 28.38 & 27.99 & 27.54 & 26.72
					\tabularnewline
					Model 3  & \ding{51} & \ding{51} & \ding{55} & \ding{55}
					& 67ms & 28.42 & 28.30 & 28.21 & 27.97 & 27.33
					\tabularnewline
					Model 4 (Ours)  & \ding{51} & \ding{51} & \ding{51} & \ding{55}
					& 70ms & 28.50 & 28.45 & 28.40 & 28.16 & 27.58
					\tabularnewline
					Model 5  & - & \ding{51} & \ding{51} & \ding{51}
					& 61ms & 28.60 & 28.67 & 28.69 & 28.48 & 27.90
					\tabularnewline
					\hline	
			\end{tabular}}
		\end{center}
		\vspace{-0.75cm}
	\end{table*}
	
	During training, 32 HR images were randomly selected and data augmentation was performed through random rotation and flipping. Then, we randomly chose 32 Gaussian kernels from the above ranges to generate LR images. For general degradations, Gaussian noises were also added to the resultant LR images. Next, 64 LR patches of size $48\times48$ (two patches from each LR image as illustrated in Sec.~\ref{sec.3.1.2}) and their corresponding HR patches were randomly cropped.  In our experiments, we set $\tau$ and $N_{queue}$ in Eq.~\ref{eq3} to 0.07 and 8192, respectively.  The Adam method \cite{Kingma2015Adam} with $\beta_{1}=0.9$ and $\beta_{2}=0.999$ was used for optimization. We first trained the degradation encoder by optimizing $L_{degrad}$ for 100 epochs. The initial learning rate was set to $1\times10^{-3}$ and decreased to $1\times10^{-4}$ after 60 epochs. Then, we trained the whole network for 500 epochs. The initial learning rate was set  $1\times10^{-4}$ and decreased to half after every 125 epochs. The overall loss function is defined as $L=L_{SR}+L_{degrad}$, where $L_{SR}$ is the $L_1$ loss between SR results and HR images.

	\subsection{Experiments on Noise-Free Degradations with Isotropic Gaussian Kernels~~~~~~~~~~~~~~~~~~~~~~~~~~~~~~~~~~~}
	\label{Sec4.2}
	
	We first conduct ablation experiments on noise-free degradations with only isotropic Gaussian kernels. Then, we compare our DASR to several recent SR networks, including RCAN \cite{Zhang2018Image}, SRMD \cite{Zhang2017Learning}, MZSR \cite{Shocher2018Zero} and IKC \cite{Gu2019Blind}. 
	RCAN is a state-of-the-art PSNR-oriented SR method for bicubic degradation. MZSR is a non-blind zero-shot SR method for degradations with isotorpic/anisotropic Gaussian kernels. SRMD is a non-blind SR method for degradations with isotropic/anisotropic Gaussian kernels and noises. IKC is a blind SR method that only considers degradations with isotropic Gaussian kernels. 
	Note that, we do not include DAN \cite{Luo2020Unfolding}, USRnet \cite{Zhang2020Deep} and correction filter \cite{Hussein2020Correction} for comparison since their degradation model is different from ours. These methods use $s$-fold downsampler\footnote{Extract upper-left pixel within each $s\times{s}$ patch.} rather than bicubic downsampler as the downsampling operation in Eq.~\ref{eq1}. To achieve fair comparison with \cite{Luo2020Unfolding,Zhang2020Deep,Hussein2020Correction}, we re-trained our DASR using their degradation model and provide the results in the supplemental material.

	\noindent \textbf{Degradation Representation Learning.}
	Degradation representation learning is used to produce discriminative representations to provide degradation information. To demonstrate its effectiveness, we introduced a network variant (Model 1) by removing degradation representation learning. Specifically, $L_{degrad}$ was excluded during training without changing the network. Besides, the separate training of degradation encoder was removed and the whole network was directly trained for 500 epochs. 

\begin{figure}
		\centering
		\includegraphics[width=1\linewidth]{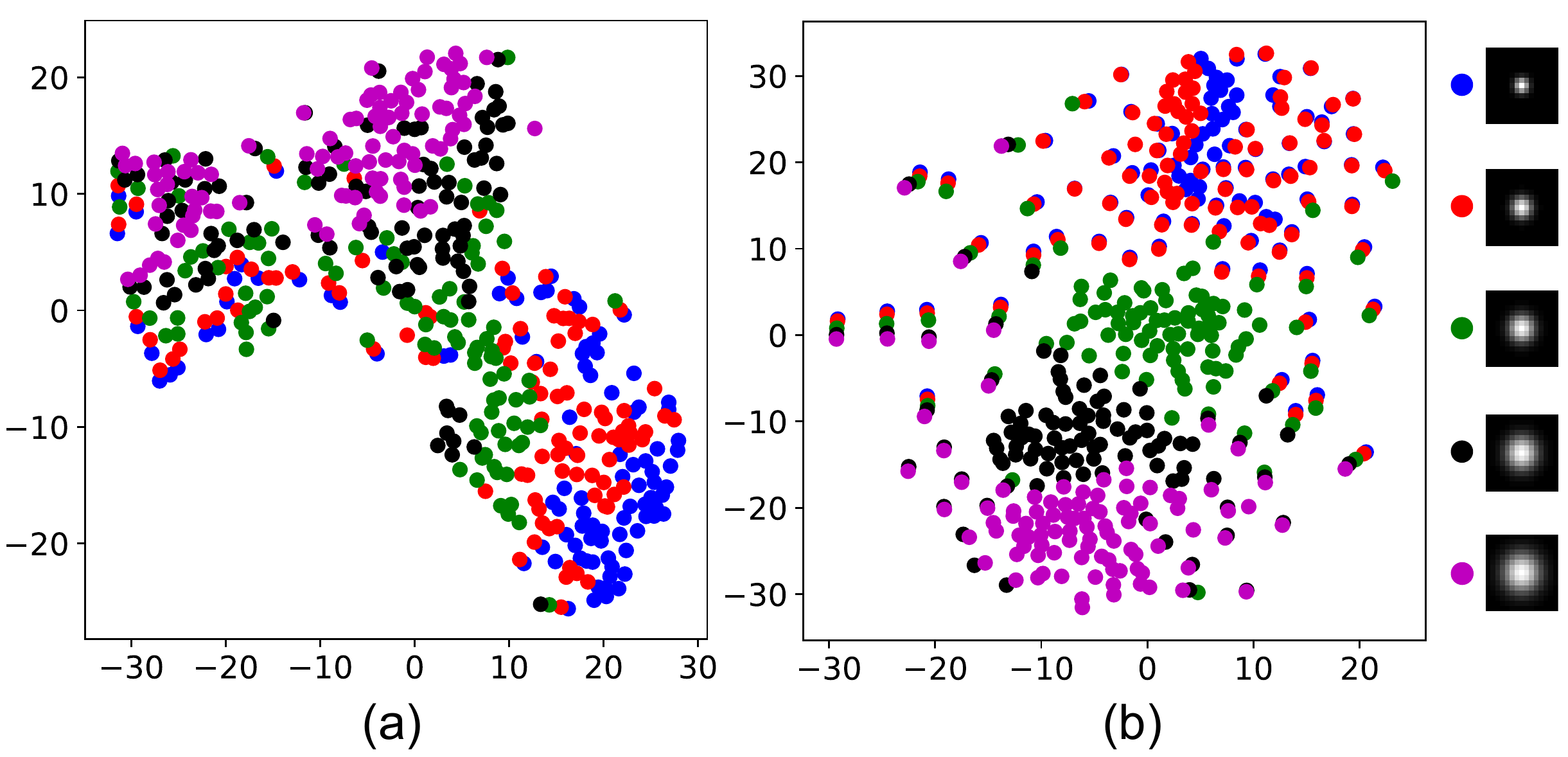}
		\caption{Visualization of representations for degradations with different kernel widths $\sigma$. (a) illustrates representations generated by our DASR w/o degradation representation learning (Model 1). (b) illustrates representations generated by our DASR (Model 4).}
		\label{fig3}
		\vspace{-0.4cm}
	\end{figure}

	We first compare the degradation representations learned by models 1 and 4. Specifically, we used B100 to generate LR images with different degradations and fed them to models 1 and 4 to produce degradation representations. Then, these representations are visualized using the T-SNE method \cite{Maaten2008Visualizing}. It can be observed in Fig.~\ref{fig3}(b) that our degradation representation learning scheme can generate discriminative clusters. Without degradation representation learning, degradations with various kernel widths cannot be well distinguished, as shown in Fig.~\ref{fig3}(a). 
	This demonstrates that degradation representation learning facilitates our degradation encoder to learn discriminative representations to provide accurate degradation information. We further compare the SR performance of models 1 and 4 in Table~\ref{tab0}. If degradation representation learning is removed, model 1 cannot handle multiple degradations well and produces lower PSNR values, especially for large kernel widths. In contrast, model 4 benefits from accurate degradation information provided by degradation representation learning to achieve better SR performance.
	
	\begin{table*}[t]
		\caption{PSNR results achieved on noise-free degradations with isotropic Gaussian kernels. \textcolor{black}{Note that, the degradation becomes bicubic degradation when kernel width is set to 0. Running time is averaged on Set14.}}
		\vspace{-0.2cm}
		\label{tab1}
		\begin{center}
			\footnotesize
			\setlength{\tabcolsep}{0.6mm}{
				\begin{tabular}{|l|c|c|cccc|cccc|cccc|cccc|}
					\hline 
					\multirow{1}{*}{Method} & \multirow{1}{*}{\tabincell{c}{Scale}}  & \multirow{1}{*}{\tabincell{c}{Time}} 
					& \multicolumn{4}{c|}{\multirow{1}{*}{Set5}} 
					& \multicolumn{4}{c|}{\multirow{1}{*}{Set14}} & \multicolumn{4}{c|}{\multirow{1}{*}{B100}}   & \multicolumn{4}{c|}{\multirow{1}{*}{Urban100}} 
					\tabularnewline
					\hline 
					\hline
					\multicolumn{3}{|l|}{\multirow{1}{*}{Kernel Width}} 
					& \textcolor{black}{0} & 0.6 & 1.2 & 1.8 
					& \textcolor{black}{0} & 0.6 & 1.2 & 1.8 
					& \textcolor{black}{0} & 0.6 & 1.2 & 1.8 
					& \textcolor{black}{0} & 0.6 & 1.2 & 1.8
					\tabularnewline
					\hline
					Bicubic 
					& \multirow{5}{*}{$\times2$}
					& - 
					& 33.66 & 32.30 & 29.28 & 27.07 
					& 30.24 & 29.21 & 27.13 & 25.47 
					& 29.56 & 28.76 & 26.93 & 25.51 
					& 26.88 & 26.13 & 24.46 & 23.06
					\tabularnewline
					RCAN \cite{Zhang2018Image} & & 170ms 
					& \textbf{38.27} & 35.91 & 31.20 & 28.50 
					& \textbf{34.12} & 32.31 & 28.48 & 26.33 
					& \textbf{32.41} & 31.16 & 28.04 & 26.26 
					& \textbf{33.34} & 29.80 & 25.38 & 23.44
					\tabularnewline
					SRMDNF \cite{Zhang2017Learning} + Predictor \cite{Gu2019Blind}
					& & 9ms 
					& 34.94 & 34.77 & 34.13 & 33.80 
					& 31.48 & 31.35 & 30.78 & 30.18 
					& 30.77 & 30.33 & 29.89 & 29.20 
					& 29.05 & 28.42 & 27.43 & 27.12
					\tabularnewline
					MZSR \cite{Soh2020Meta} + Predictor \cite{Gu2019Blind}
					& & 90ms 
					& 35.96 & 35.66 & 35.22 & 32.32 
					& 31.97 & 31.33 & 30.85 & 29.17 
					& 30.64 & 29.82 & 29.41 & 28.72 
					& 29.49 & 29.01 & 28.43 & 26.39
					\tabularnewline
					DASR (Ours)  
					&  & 71ms 
					& 37.87 & \textbf{37.47} & \textbf{37.19} & \textbf{35.43} 
					& 33.34 & \textbf{32.96} & \textbf{32.78} & \textbf{31.60} 
					& 32.03 & \textbf{31.78} & \textbf{31.71} & \textbf{30.54} 
					& 31.49 & \textbf{30.71} & \textbf{30.36} & \textbf{28.95}
					\tabularnewline
					\hline
					\hline
					\multicolumn{3}{|l|}{Kernel Width} 
					& 0 & 0.8 & 1.6 & 2.4 
					& 0 & 0.8 & 1.6 & 2.4  
					& 0 & 0.8 & 1.6 & 2.4  
					& 0 & 0.8 & 1.6 & 2.4 
					\tabularnewline
					\hline
					Bicubic
					& \multirow{4}{*}{$\times3$}
					& - 
					& 30.39 & 29.42 & 27.24 & 25.37 
					& 27.55 & 26.84 & 25.42 & 24.09 
					& 27.21 & 26.72 & 25.52 & 24.41 
					& 24.46 & 24.02 & 22.95 & 21.89
					\tabularnewline
					RCAN \cite{Zhang2018Image} & & 167ms 
					& \textbf{34.74} & 32.90 & 29.12 & 26.75 
					& \textbf{30.65} & 29.49 & 26.75 & 24.99 
					& \textbf{29.32} & 28.56 & 26.55 & 25.18 
					& \textbf{29.09} & 26.89 & 23.85 & 22.30
					\tabularnewline
					SRMDNF \cite{Zhang2017Learning} + Predictor \cite{Gu2019Blind}
					& & 6ms 
					& 32.22 & 32.63 & 32.27 & 28.62 
					& 29.13 & 29.25 & 28.01 & 26.90 
					& 28.41 & 28.25 & 28.11 & 26.56 
					& 26.75 & 26.61 & 26.35 & 24.06
					\tabularnewline
					DASR (Ours) 
					& & 70ms 
					& 34.11 & \textbf{34.08} & \textbf{33.57} & \textbf{32.15} 
					& 30.13 & \textbf{29.99} & \textbf{28.66} & \textbf{28.42} 
					& 28.96 & \textbf{28.90} & \textbf{28.62} & \textbf{28.13} 
					& 27.65 & \textbf{27.36} & \textbf{26.86} & \textbf{25.95}
					\tabularnewline				
					\hline
					\hline
					\multicolumn{3}{|l|}{Kernel Width} 
					& 0 & 1.2 & 2.4 & 3.6 
					& 0 & 1.2 & 2.4 & 3.6 
					& 0 & 1.2 & 2.4 & 3.6 
					& 0 & 1.2 & 2.4 & 3.6
					\tabularnewline
					\hline
					Bicubic 
					& \multirow{5}{*}{$\times4$}
					& - 
					& 28.42 & 27.30 & 25.12 & 23.40 
					& 26.00 & 25.24 & 23.83 & 22.57 
					& 25.96 & 25.42 & 24.20 & 23.15 
					& 23.14 & 22.68 & 21.62 & 20.65
					\tabularnewline
					RCAN \cite{Zhang2018Image} & & 165ms 
					& \textbf{32.63} & 30.26 & 26.72 & 24.66 
					& \textbf{28.87} & 27.48 & 24.93 & 23.41 
					& \textbf{27.72} & 26.89 & 25.09 & 23.93 
					& \textbf{26.61} & 24.71 & 22.25 & 20.99
					\tabularnewline
					SRMDNF \cite{Zhang2017Learning} + Predictor \cite{Gu2019Blind}
					& & 5ms 
					& 30.61 & 29.35 & 29.27 & 28.65 
					& 27.74 & 26.15 & 26.20 & 26.17 
					& 27.15 & 26.15 & 26.15 & 26.14 
					& 25.06 & 24.11 & 24.10 & 24.08
					\tabularnewline
					IKC \cite{Gu2019Blind} 
					& & 517ms 
					& 32.00 & 31.77 & 30.56 & 29.23 
					& 28.52 & 28.45 & 28.16 & 26.81 
					& 27.51 & 27.43 & 27.27 & 26.33 
					& 25.93 & 25.63 & 25.00 & 24.06
					\tabularnewline
					DASR (Ours) 
					& & 70ms 
					& 31.99 & \textbf{31.92} & \textbf{31.75} & \textbf{30.59} 
					& 28.50 & \textbf{28.45} & \textbf{28.28} & \textbf{27.45} 
					& 27.51 & \textbf{27.52} & \textbf{27.43} & \textbf{26.83} 
					& 25.82 & \textbf{25.69} & \textbf{25.44} & \textbf{24.66}
					\tabularnewline	
					\hline	
			\end{tabular}}
		\end{center}
		\vspace{-0.7cm}
	\end{table*}

	\noindent \textbf{Degradation-Aware Convolutions.}
	With degradation encoder, the extracted degradation representation is incorporated by DA convolutions to achieve flexible adaption to different degradations by predicting convolutional kernels and channel-wise modulation coefficients. To demonstrate the effectiveness of these two key components, we first introduced a variant (Model 2) by replacing DA convolutions with vanilla ones. Specifically, degradation representations are stretched and concatenated with image features as in \cite{Zhang2017Learning} before being fed to vanilla convolutions. 
	Then, we developed another variant (Model 3) by removing the channel-wise modulation coefficient branch.
	Note that, we adjust the number of channels in models 2 and 3 to ensure comparable model sizes. 
	From Table~\ref{tab0} we can see that our DASR benefits from both dynamic convolutional kernels and channel-wise modulation coefficients to produce better results for various degradations.

	\noindent \textbf{Blind SR vs. Non-Blind SR.}
	We further investigate the upper-bound performance of our DASR network by providing groundtruth degradation. Specifically, we replaced the degradation encoder with 5 FC layers to learn a representation directly from the true degradation (\emph{i.e.}, blur kernel). This network variant (Model 5) was then trained from scratch for 500 epochs. When groundtruth degradation is provided, model 5 achieves improved performance and outperforms SRMDNF by notable margins. Further, SRMDNF is quite sensitive to degradation estimation errors under blind settings, with PSNR values being decreased if degradation is not accurately estimated (\emph{e.g.}, 27.55 vs. 26.66/26.18 for $\sigma\!=\!3.4$). In contrast, our DASR (Model 4) benefits from degradation representation learning to achieve better blind SR performance.
	
\begin{figure}
		\centering
		\includegraphics[width=1\linewidth]{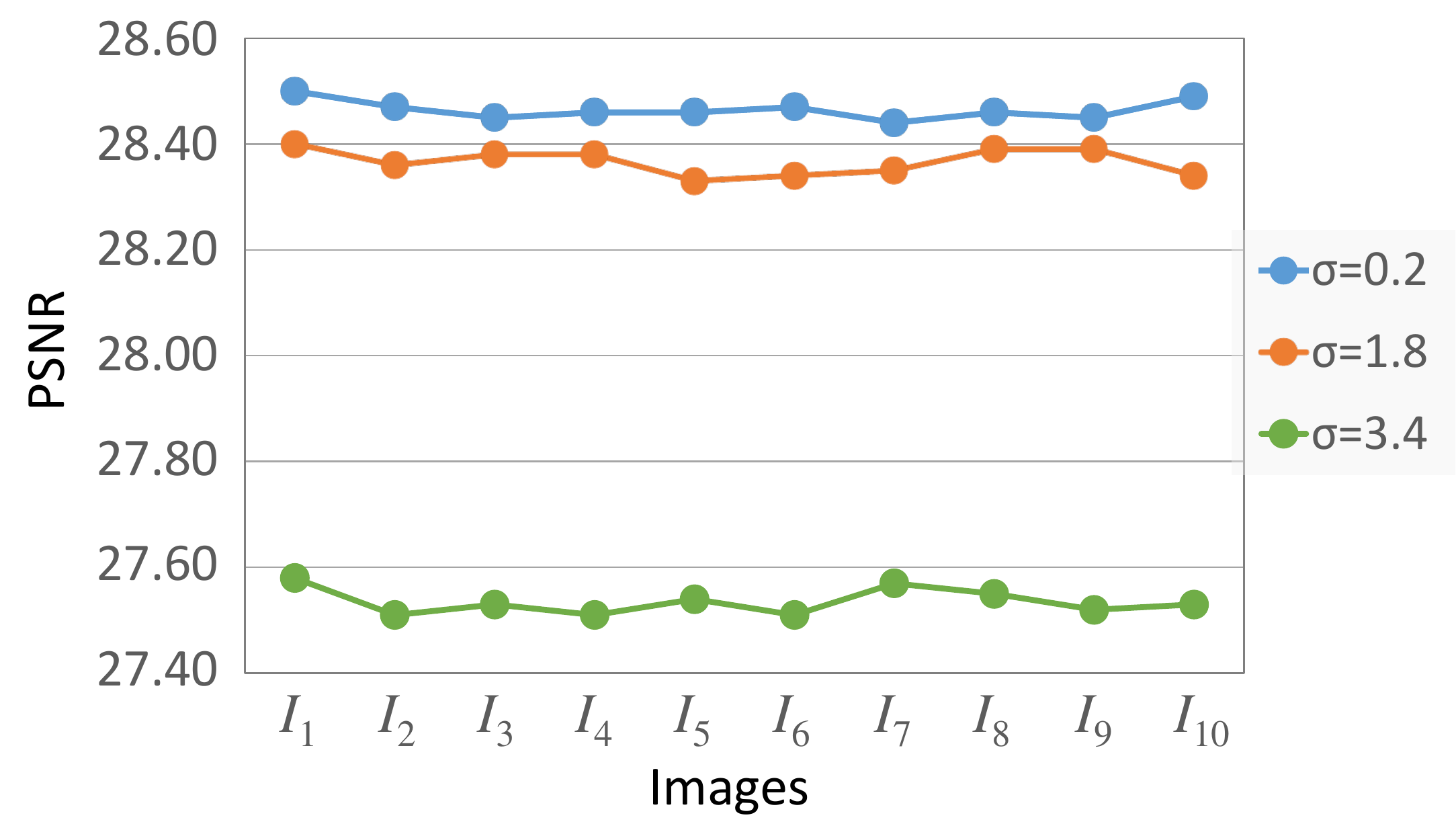}
		\caption{PSNR results achieved on Set14 using degradation representations learned from different image contents.}
		\label{fig8}
		\vspace{-0.35cm}
	\end{figure}

	\noindent \textbf{Study of Degradation Representations.} 
	Our degradation representations aim at extracting content-invariant degradation information from LR images. 
	To demonstrate this, we conduct experiments to study the effect of different image contents to our degradation representations. Specifically, given an HR image, we first generated an LR image $I_1$ using a Gaussian kernel $k$. Then, we randomly selected another 9 HR images to generate LR images ($I_i (i=2,3,...10)$) using $k$. Next, degradation representations were extracted from $I_i (i=1,2,...10)$ to super-resolve $I_1$. Note that, $I_i (i=2,3,...10)$ and $I_1$ share the same degradation but have different image contents. 
	From Fig.~\ref{fig8} we can see that our network achieves relatively stable performance with degradation representations learned from different image contents. This demonstrates that our degradation representations are robust to image content variations.

	\begin{figure*}
		\centering
		\includegraphics[width=1\linewidth]{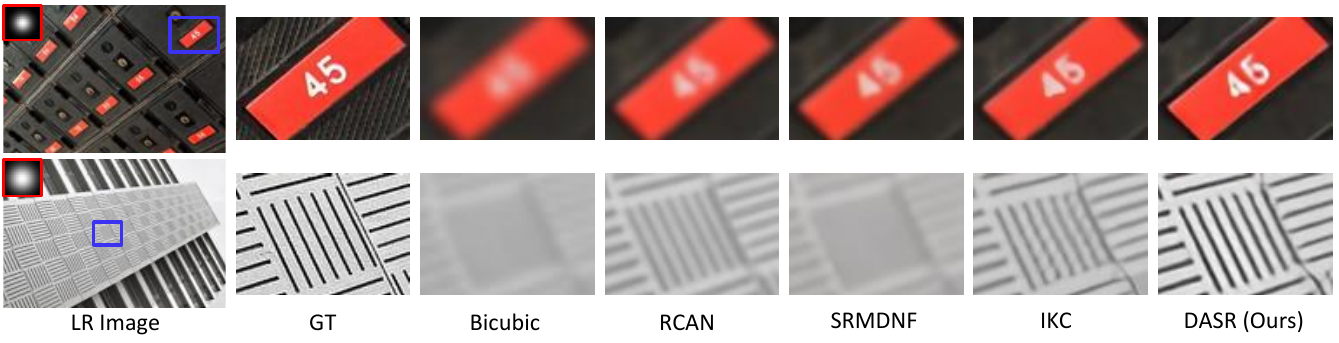}
		\vspace{-0.55cm}
		\caption{Visual comparison achieved on Urban100 for $\times4$ SR. The blur kernels are illustrated with red boxes.}
		\label{fig5}
		\vspace{-0.4cm}
	\end{figure*}
	
	\noindent \textbf{Comparison to Previous Networks.}
	We compare DASR to RCAN, SRMD, MZSR and IKC. Pre-trained models of these networks are used for evaluation following their default settings. Quantitative results are shown in Table~\ref{tab1}, while visualization results are provided in Fig.~\ref{fig5}. Note that, MZSR\footnote{Pre-trained $\times4$ model of MZSR is based on $s$-fold downsampler.}/IKC are only tested for $\times2/4$ SR since their pre-trained models for other scale factors are unavailable. For non-blind SR methods (SRMD and MZSR), we first performed degradation estimation to provide degradation information. Since KernelGAN is quite time-consuming (Table~\ref{tab0}), the predictor sub-network in IKC was used to estimate degradations. 
	
	\textcolor{black}{It can be observed from Table~\ref{tab1} that RCAN produces the highest PSNR results on bicubic degradation (\emph{i.e.}, kernel width 0) while suffering relatively low performance when the test degradations are different from the bicubic one.} Although SRMDNF and MZSR can adapt to the estimated degradations, these methods are sensitive to degradation estimation, as demonstrated in Table~\ref{tab0}. Therefore, degradation estimation errors can be magnified by SRMDNF and MZSR, resulting in limited SR performance. 
	Since an iterative correction scheme is used to correct the estimated degradation, IKC outperforms SRMDNF with higher PSNR values being achieved. However, IKC is time-consuming due to its iterations. Compared to IKC, our DASR network achieves better performance for different degradations with shorter running time. That is because, our degradation representation learning scheme can extract ``good'' representations to distinguish different degradations at a single inference. 
	
	Visualization results achieved by different methods are shown in 	Fig.~\ref{fig5}. Since RCAN is trained on the fixed bicubic degradation, it cannot reliably recover missing details when the real degradation differs from the bicubic one. Although SRMDNF can handle multiple degradations, failures can be caused by the degradation estimation error. By iteratively correcting the estimated degradations, IKC achieves better performance than SRMDNF. Compared to other methods, our DASR produces results with much clearer details and higher perceptual quality.

	\subsection{Experiments on General Degradations with Anisotropic Gaussian Kernels and Noises}
	We further conduct experiments on general degradations with anisotropic Gaussian kernels and noises. We first analyze the representations learned from general degradations and then compare the performance of our DASR to RCAN, SRMDNF and IKC under blind settings.

\begin{figure}
		\centering
		\includegraphics[width=1\linewidth]{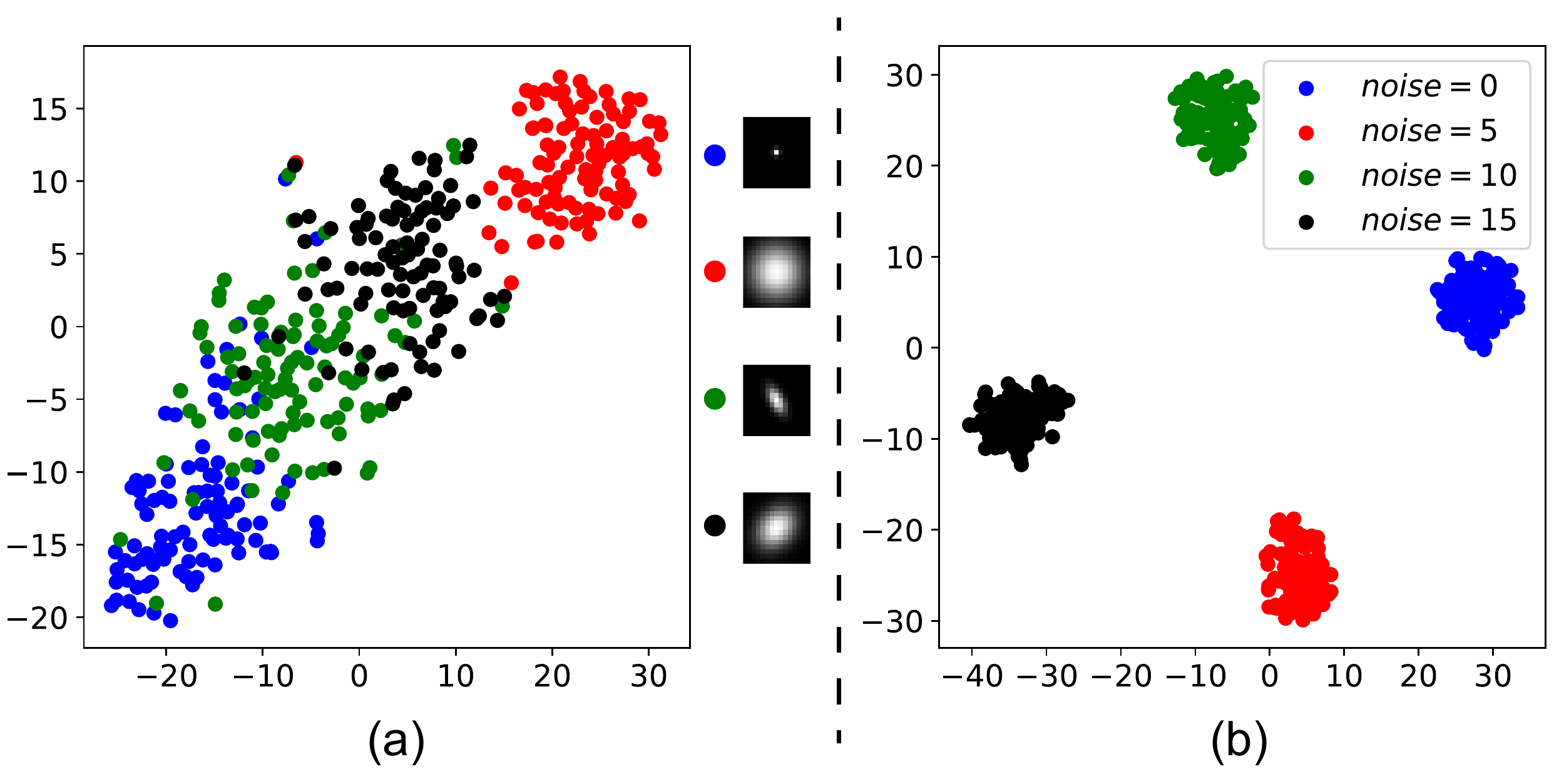}
		\caption{Visualization of representations for degradations with different blur kernels (a) and noise levels (b).}
		\label{fig9}
		\vspace{-0.3cm}
	\end{figure}

	\noindent \textbf{Study of Degradation Representations.}
	Experiments are conducted to investigate the effect of two different components (\emph{i.e.}, blur kernels and noises) to our degradation representations. We first visualize the representations for noise-free degradations with various blur kernels in Fig.~\ref{fig9}(a). Then, we randomly select a blur kernel and visualize the representations for degradations with different noise levels in Fig.~\ref{fig9}(b). It can be observed that our degradation encoder can easily cluster degradations with different noise levels into discriminative groups and roughly distinguish various blur kernels. 
	
	\begin{table*}[t]
		\caption{PSNR results achieved on Set14 for $\times4$ SR.}
		\label{tab2}
		\begin{center}
			\footnotesize
			\setlength{\tabcolsep}{1mm}{
				\begin{tabular}{|l|l|c|c|ccccccccc|}
					\hline 
					\multirow{2}{*}{\tabincell{c}{\\Method}}
					& \multirow{2}{*}{\tabincell{c}{\\\#Params.}}
					& \multirow{2}{*}{\tabincell{c}{\\Time}}
					& \multirow{2}{*}{\tabincell{c}{\\Noise}}
					& \multicolumn{9}{c|}{Blur Kernel}
					\tabularnewline
					& & & 
					& \begin{minipage}[b]{0.07\columnwidth}
						\centering
						\raisebox{-.5\height}{\includegraphics[width=\linewidth]{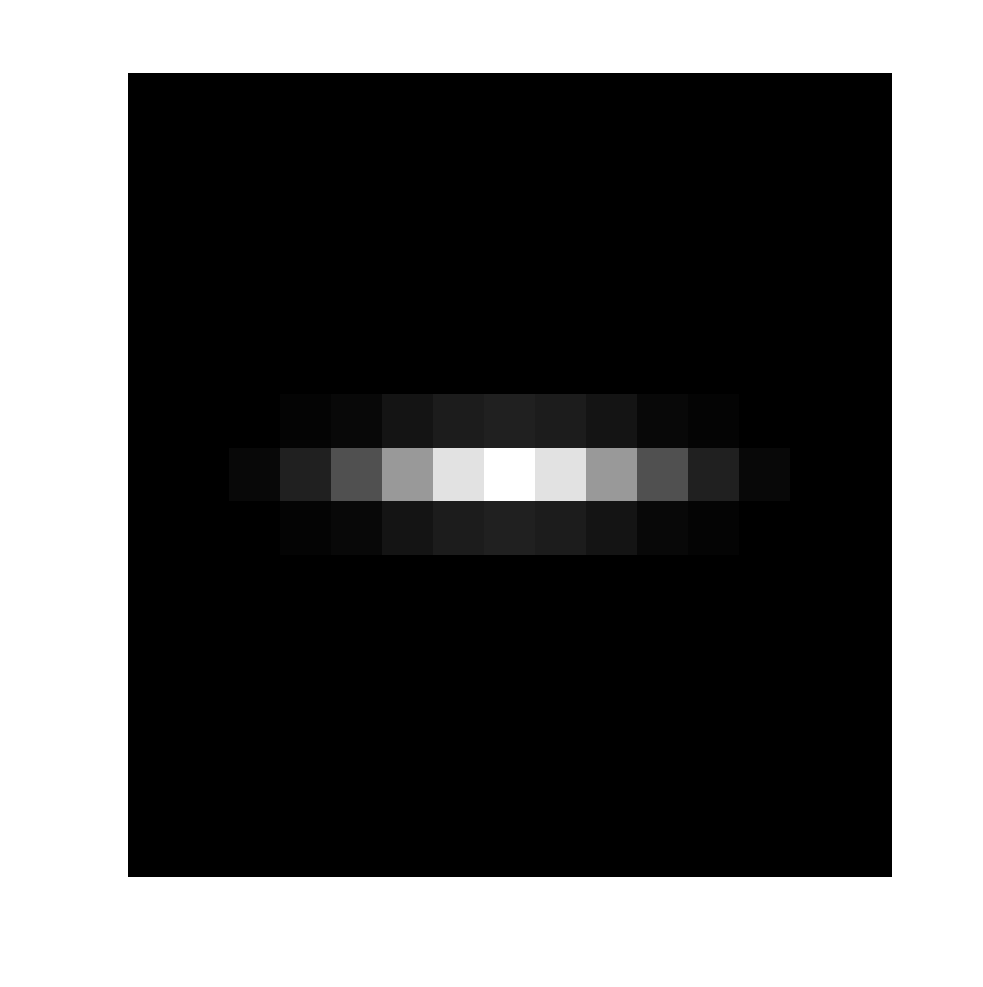}}
					\end{minipage}
					& \begin{minipage}[b]{0.07\columnwidth}
						\centering
						\raisebox{-.5\height}{\includegraphics[width=\linewidth]{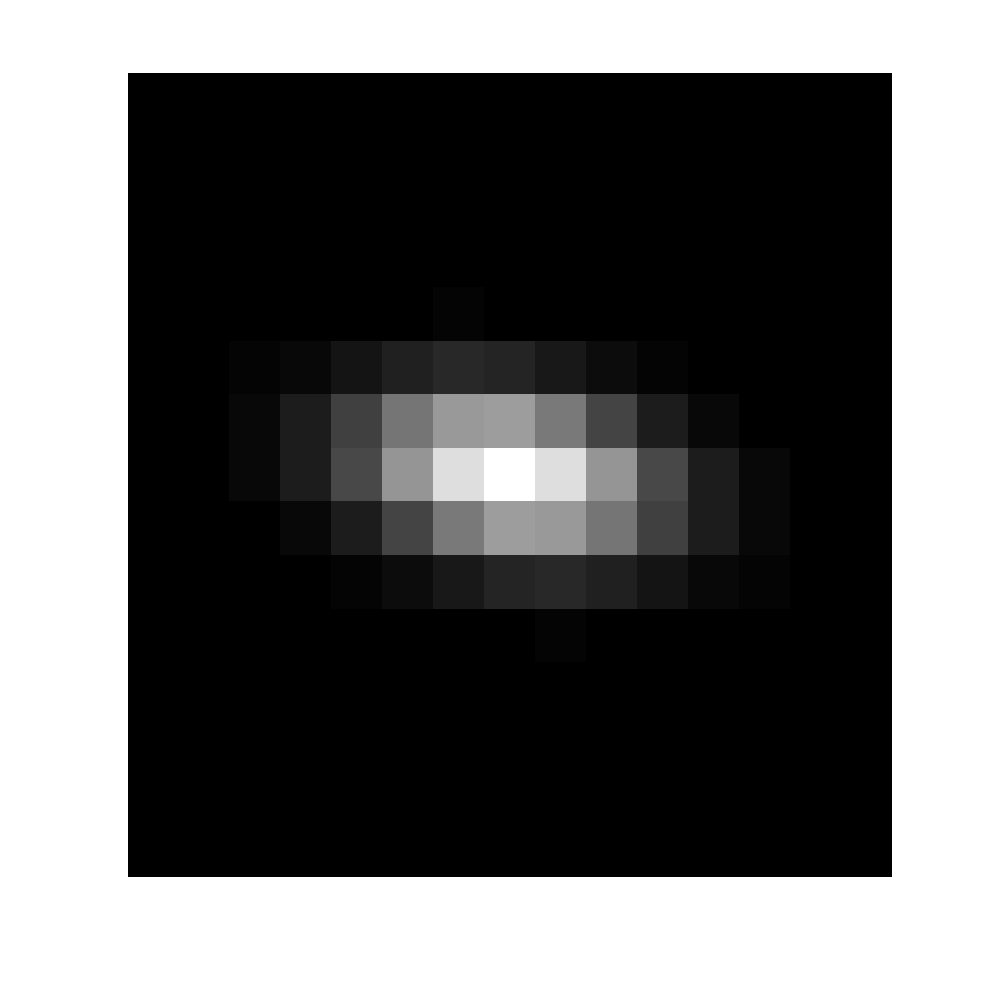}}
					\end{minipage}
					& \begin{minipage}[b]{0.07\columnwidth}
						\centering
						\raisebox{-.5\height}{\includegraphics[width=\linewidth]{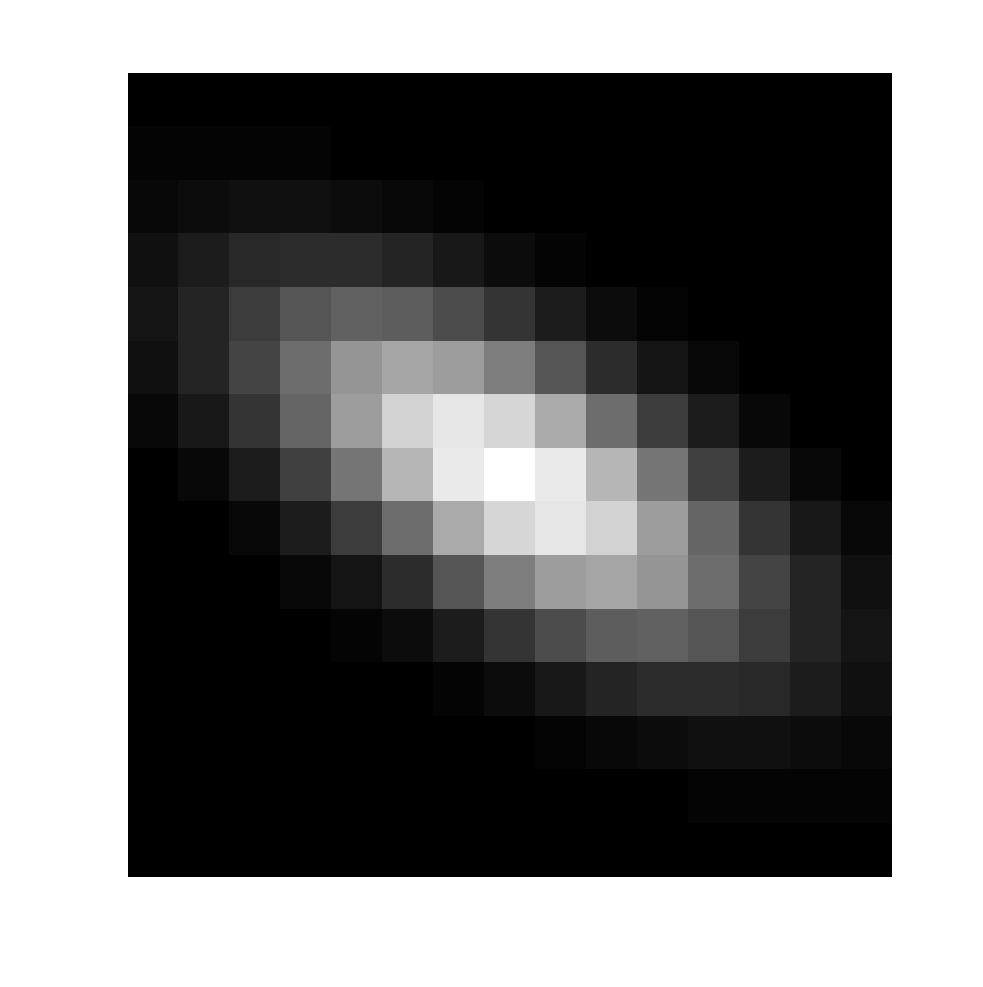}}
					\end{minipage}
					& \begin{minipage}[b]{0.07\columnwidth}
						\centering
						\raisebox{-.5\height}{\includegraphics[width=\linewidth]{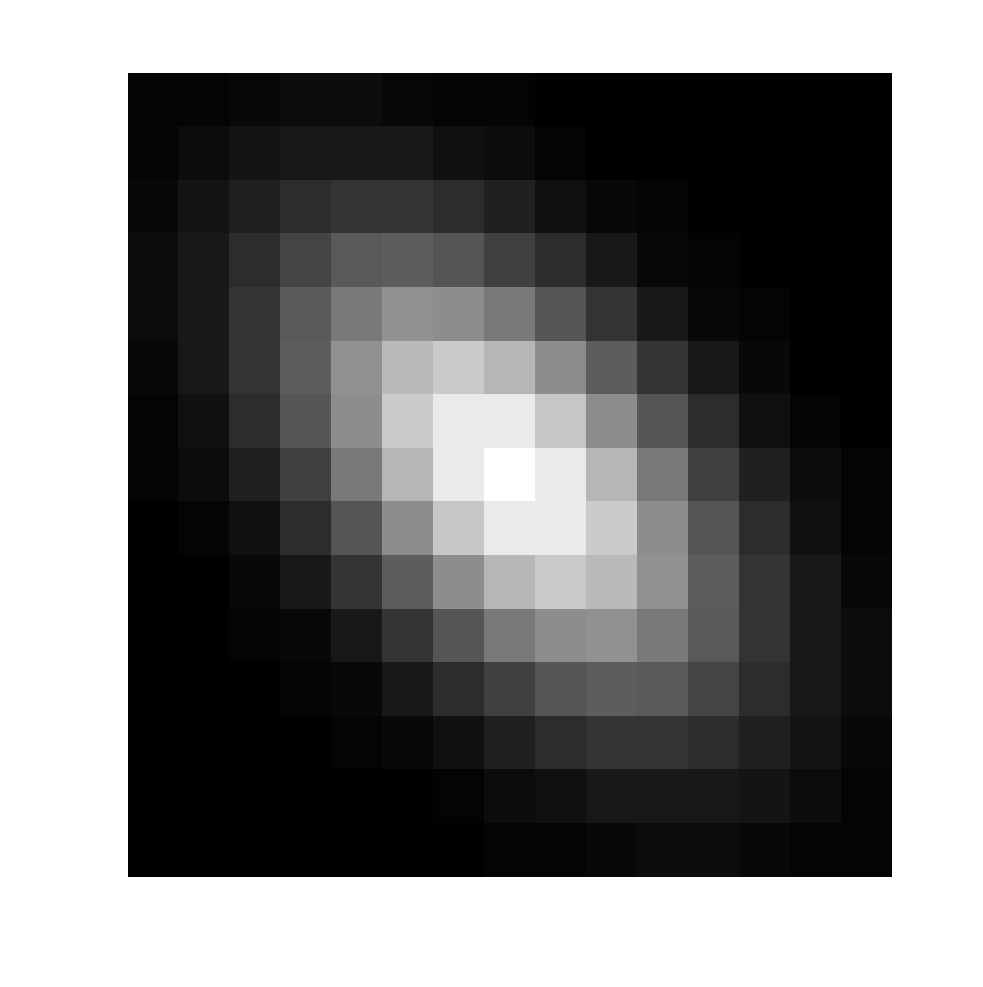}}
					\end{minipage}
					& \begin{minipage}[b]{0.07\columnwidth}
						\centering
						\raisebox{-.5\height}{\includegraphics[width=\linewidth]{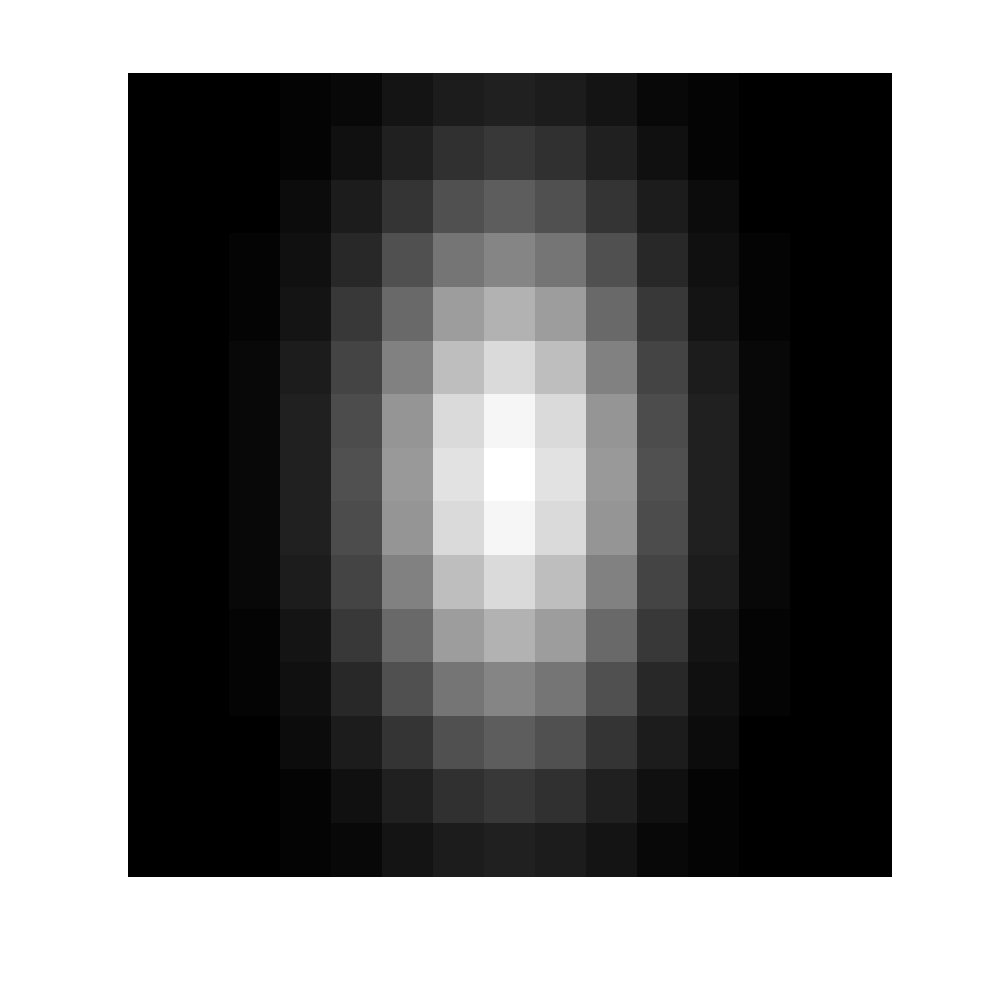}}
					\end{minipage}
					& \begin{minipage}[b]{0.07\columnwidth}
						\centering
						\raisebox{-.5\height}{\includegraphics[width=\linewidth]{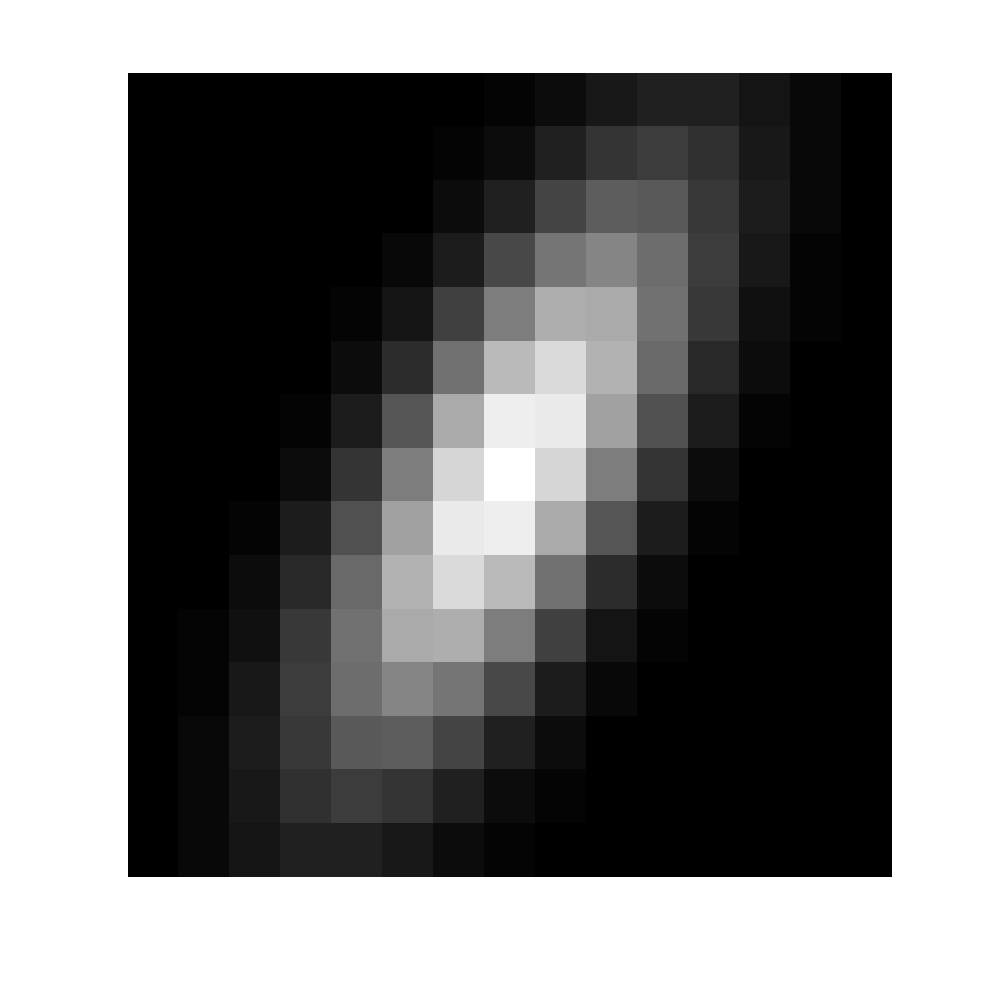}}
					\end{minipage}
					& \begin{minipage}[b]{0.07\columnwidth}
						\centering
						\raisebox{-.5\height}{\includegraphics[width=\linewidth]{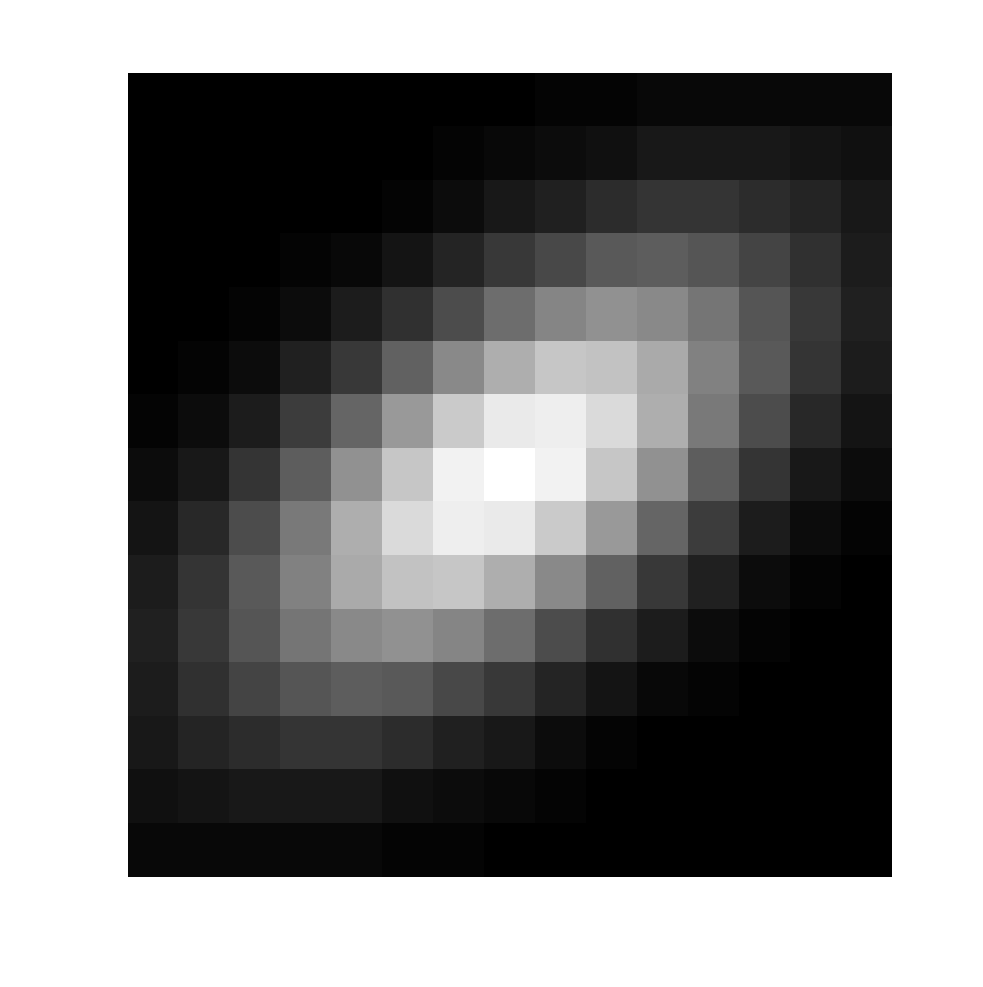}}
					\end{minipage}
					& \begin{minipage}[b]{0.07\columnwidth}
						\centering
						\raisebox{-.5\height}{\includegraphics[width=\linewidth]{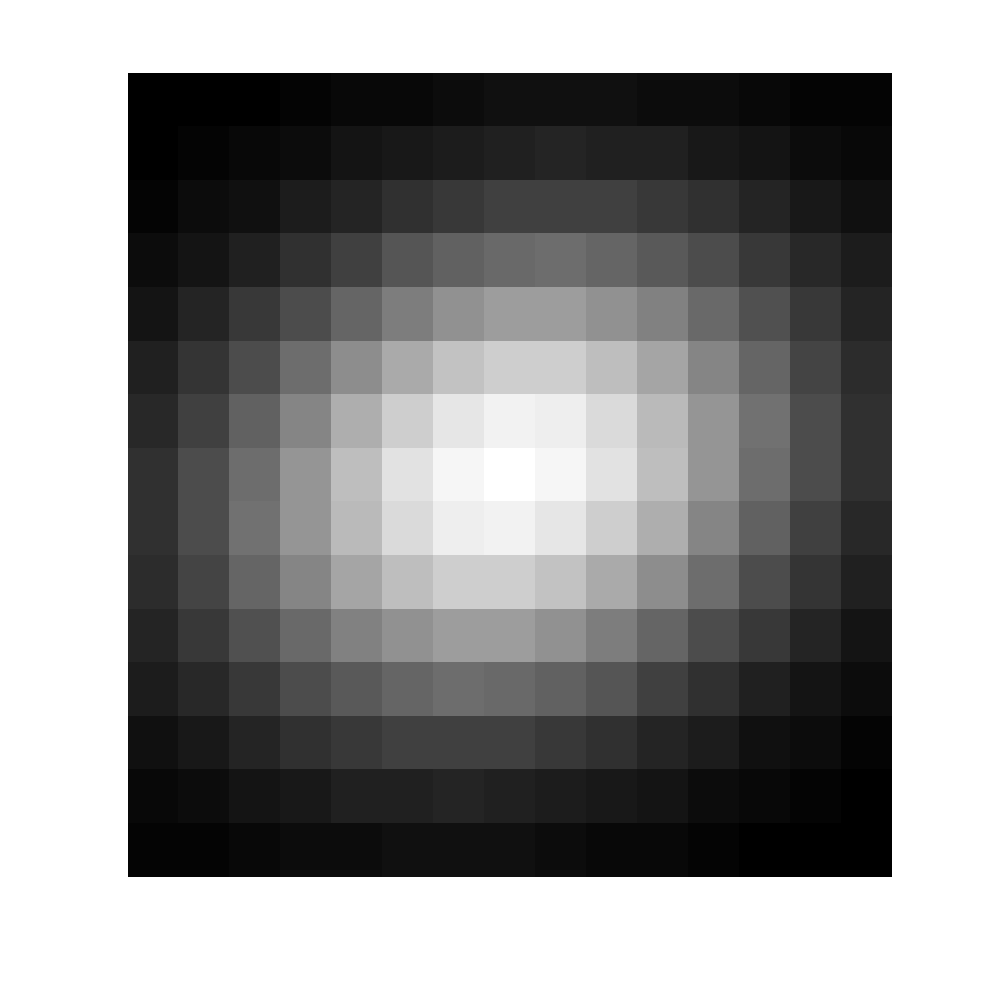}}
					\end{minipage}
					& \begin{minipage}[b]{0.07\columnwidth}
						\centering
						\raisebox{-.5\height}{\includegraphics[width=\linewidth]{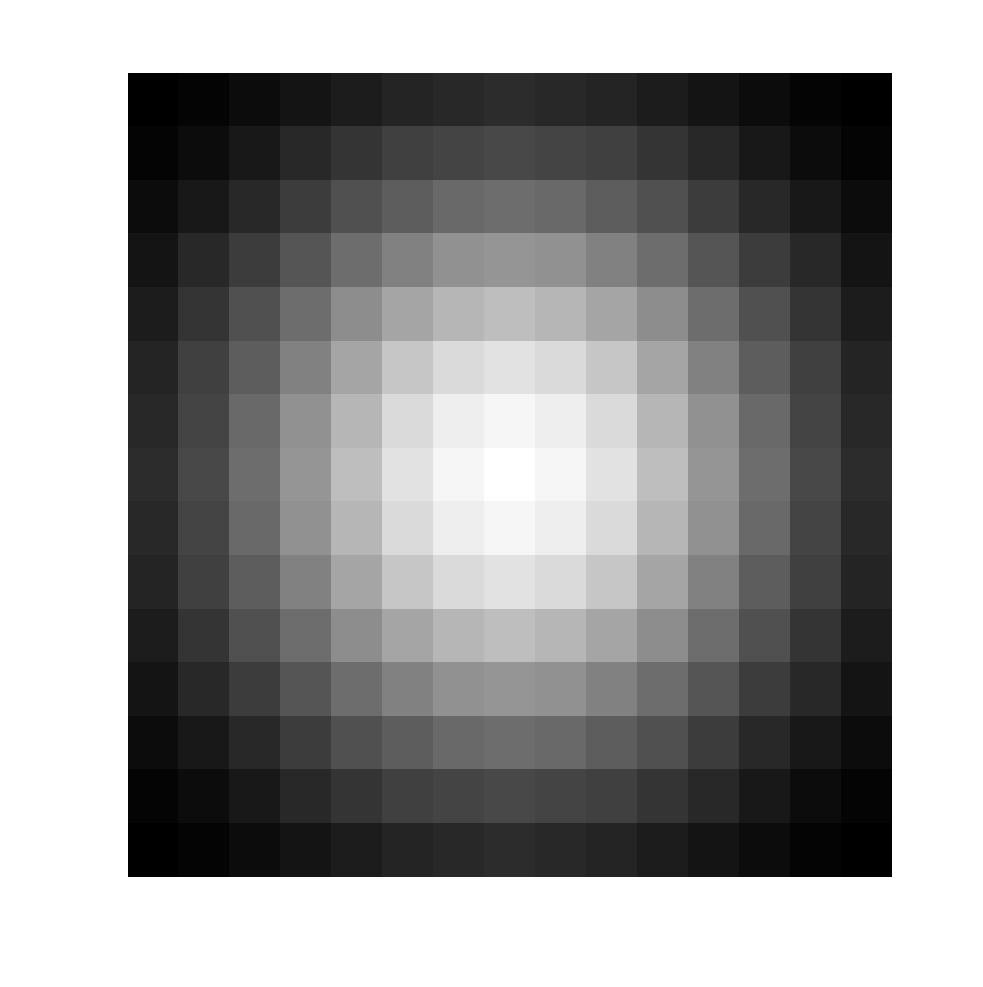}}
					\end{minipage}
					\tabularnewline
					\hline
					\hline
					\multirow{3}{*}{\tabincell{l}{DnCNN \cite{Zhang2017Gaussian}+RCAN \cite{Zhang2018Image}}}
					& \multirow{3}{*}{650K+15.2M} & \multirow{3}{*}{169ms} & 0 & 26.44 & 26.22 & 24.48 & 24.23 & 24.29 & 24.19 & 23.90 & 23.42 & 23.01
					\tabularnewline
					&  & & 5 & 26.10 & 25.90 & 24.29 & 24.07 & 24.14 & 24.02 & 23.74 & 23.31 & 22.92
					\tabularnewline
					&  & & 10 & 25.65 & 25.47 & 24.05 & 23.84 & 23.92 & 23.80 & 23.54 & 23.14 & 22.77
					\tabularnewline
					\hline
					\multirow{3}{*}{\tabincell{l}{DnCNN \cite{Zhang2017Gaussian}+SRMDNF \cite{Zhang2018Image}+Predictor \cite{Gu2019Blind}}}
					& \multirow{3}{*}{650K+1.5M+420K} & \multirow{3}{*}{8ms} & 0 & 26.84 & 26.88 & 25.57 & 25.69 & 25.64 & 24.98 & 25.12 & 25.28 & 24.84
					\tabularnewline
					&  & & 5 & 25.92 & 25.75 & 24.18 & 23.97 & 24.05 & 23.93 & 23.65 & 23.20 & 22.80
					\tabularnewline
					&  & & 10 & 25.39 & 25.23 & 23.88 & 23.68 & 23.74 & 23.65 & 23.39 & 22.99 & 22.64
					\tabularnewline
					\hline
					\multirow{3}{*}{\tabincell{l}{DnCNN \cite{Zhang2017Gaussian}+IKC \cite{Gu2019Blind}}}	
					& \multirow{3}{*}{650K+5.2M} & \multirow{3}{*}{520ms} & 0 & 27.71 & 27.78 & 27.11 & 27.02 & 26.93 & 26.65 & 26.50 & 26.01 & 25.33
					\tabularnewline
					&  & & 5 & 26.91 & 26.80 & 24.87 & 24.53 & 24.56 & 24.40 & 24.06 & 23.53 & 23.06
					\tabularnewline
					&  & & 10 & 26.16 & 26.09 & 24.55 & 24.33 & 24.35 & 24.17 & 23.92 & 23.43 & 23.01
					\tabularnewline
					\hline
					\multirow{3}{*}{DASR (Ours)}
					& \multirow{3}{*}{5.8M} & \multirow{3}{*}{70ms} & 0 & \textbf{27.99} & \textbf{27.97} & \textbf{27.53} & \textbf{27.45} & \textbf{27.43} & \textbf{27.22} & \textbf{27.19} & \textbf{26.83} & \textbf{26.21}
					\tabularnewline
					&  & & 5 & \textbf{27.25} & \textbf{27.18} & \textbf{26.37} & \textbf{26.16} & \textbf{26.09} & \textbf{25.96} & \textbf{25.85} & \textbf{25.52} & \textbf{25.04}
					\tabularnewline
					&  & & 10 & \textbf{26.57} & \textbf{26.51} & \textbf{25.64 }& \textbf{25.47} & \textbf{25.43} & \textbf{25.31} & \textbf{25.16} & \textbf{24.80} & \textbf{24.43}
					\tabularnewline
					\hline
			\end{tabular}}
		\end{center}
		\vspace{-0.6cm}
	\end{table*}
	
	\begin{figure*}
		\centering
		\includegraphics[width=0.9\linewidth]{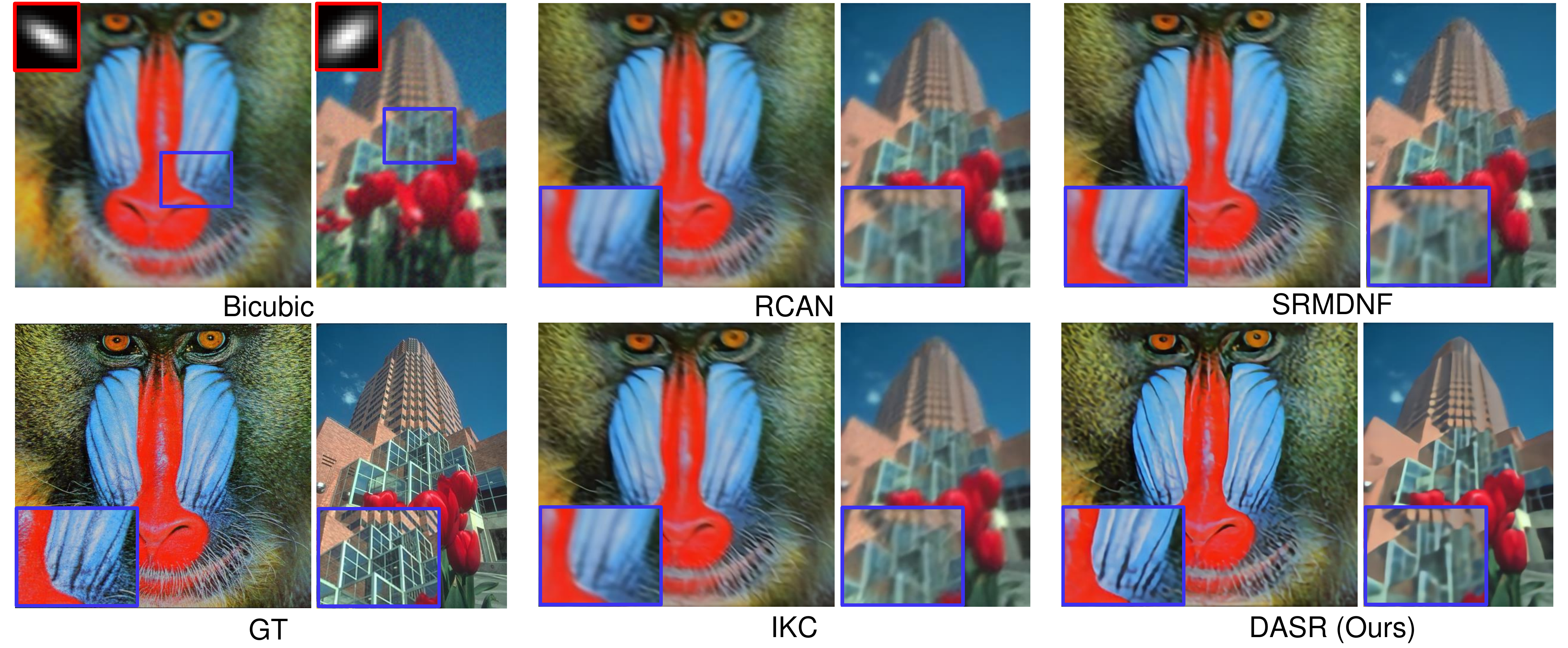}
		\vspace{-0.2cm}
		\caption{Visual comparison achieved on Set14 and B100. Noise levels are set to 0 and 5 for these two images, respectively.}
		\label{fig6}
		\vspace{-0.3cm}
	\end{figure*}
	
	\noindent \textbf{Comparison to Previous Networks.}
	We use 9 typical blur kernels and different noise levels for performance evaluation. To super-resolve noisy LR images using RCAN, SRMDNF and IKC, we first denoise the LR images using DnCNN \cite{Zhang2017Gaussian} (a state-of-the-art denoising method) under blind settings. 
	Since the pre-trained model of IKC is trained on isotropic Gaussian kernels only, we further fine-tuned this model on anisotropic Gaussian kernels for fair comparison. 
	The predictor sub-network of the fine-tuned IKC model is used to estimate degradations for SRMDNF.
	
	It can be observed from Table~\ref{tab2} that RCAN produces relatively low performance on complex degradations since it is trained on bicubic degradation only. Since SRMDNF is sensitive to degradation estimation errors, its performance for complex degradations is limited. By iteratively correcting the estimated degradations, IKC performs favorably against SRMDNF. However, IKC is more time-consuming since numerous iterations are required. Different from IKC that focuses on pixel-level degradation estimation, our DASR explores an effective yet efficient approach to learn discriminative representations to distinguish different degradations. Using our degradation representation learning scheme, DASR outperforms IKC in terms of PSNR for various blur kernels and noise levels with running time being reduced by over 7 times. 
	Figure~\ref{fig6} further illustrates the visualization results produced by different methods. Our DASR achieves much better visual quality while other methods suffer obvious blurring artifacts.

	\subsection{Experiments on Real Degradations}
\begin{figure}
		\centering
		\includegraphics[width=0.92\linewidth]{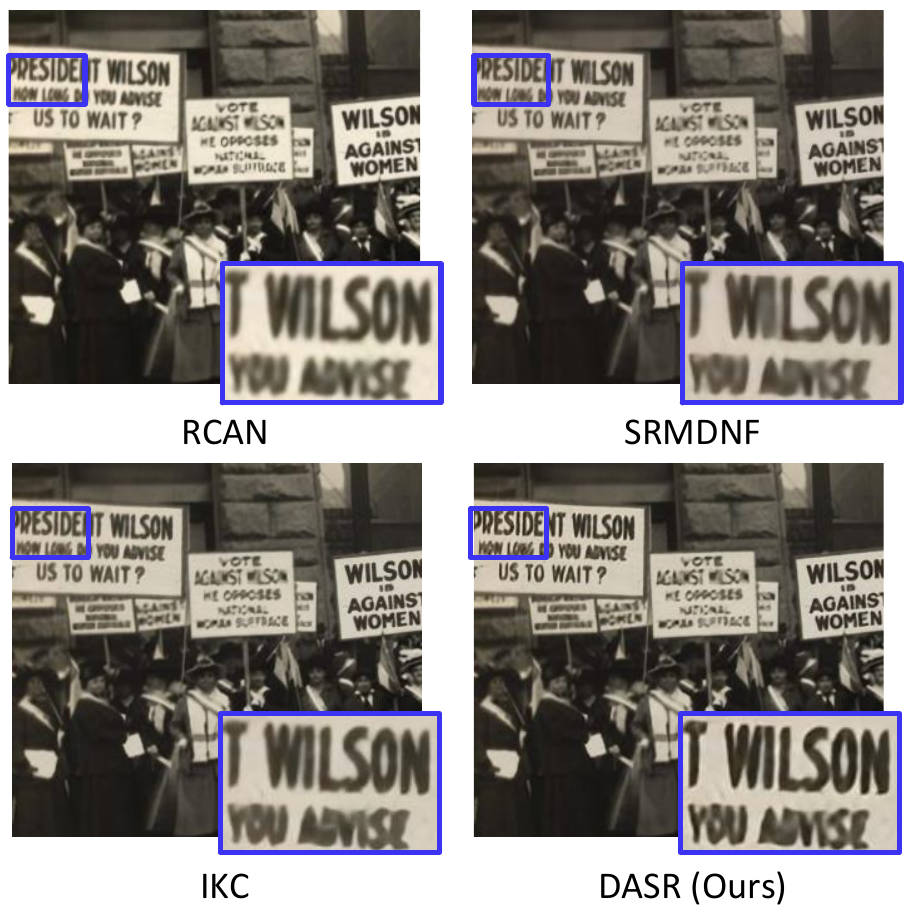}
		\caption{Visualization results achieved on a real image.}
		\label{fig7}
		\vspace{-0.3cm}
	\end{figure}

	We further conduct experiments on real degradations to demonstrate the effectiveness of our DASR. Following \cite{Zhang2017Learning}, DASR trained on isotropic Gaussian kernels is used for evaluation on real images. Visualization results are shown in Fig.~\ref{fig7}. It can be observed that our DASR produces visually more promising results with clearer details and fewer blurring artifacts.

	\section{Conclusion}
	In this paper, we proposed an unsupervised degradation representation learning scheme for blind SR to handle various degradations. Instead of explicitly estimating the degradations, we use contrastive learning to extract discriminative representations to distinguish different degradations. Moreover, we introduce a degradation-aware SR (DASR) network with flexible adaption to different degradations based on the learned representations. It is demonstrated that our degradation representation learning scheme can extract discriminative representations to obtain accurate degradation information. Experimental results show that our network achieves state-of-the-art performance for blind SR with various degradations.

	\section*{Acknowledge}
	The authors would like to thank anonymous reviewers for their insightful suggestions. Xiaoyu Dong is supported by RIKEN Junior Research Associate Program. Part of this work was done when she was a master student at HEU.

	{\small
		\bibliographystyle{ieee_fullname}
		\bibliographystyle{unsrt}
		\bibliography{super-resolution,other-CV-fields,neural-network}

\begin{thebibliography}{10}\itemsep=-1pt

\bibitem{Agustsson2017NTIRE}
Eirikur Agustsson and Radu Timofte.
\newblock {NTIRE} 2017 challenge on single image super-resolution: Dataset and
  study.
\newblock In {\em CVPRW}, pages 1122--1131, 2017.

\bibitem{Ahn2018Fast}
Namhyuk Ahn, Byungkon Kang, and Kyung{-}Ah Sohn.
\newblock Fast, accurate, and lightweight super-resolution with cascading
  residual network.
\newblock In {\em ECCV}, pages 252--268, 2018.

\bibitem{Bell-Kligler2019Blind}
Sefi Bell-Kligler, Assaf Shocher, and Michal Irani.
\newblock Blind super-resolution kernel estimation using an internal-gan.
\newblock In {\em NeurIPS}, pages 284--293, 2019.

\bibitem{Bevilacqua2012Low}
Marco Bevilacqua, Aline Roumy, Christine Guillemot, and Marie{-}Line
  Alberi{-}Morel.
\newblock Low-complexity single-image super-resolution based on nonnegative
  neighbor embedding.
\newblock In {\em BMVC}, pages 1--10, 2012.

\bibitem{Chen2020simple}
Ting Chen, Simon Kornblith, Mohammad Norouzi, and Geoffrey Hinton.
\newblock A simple framework for contrastive learning of visual
  representations.
\newblock In {\em ICML}, 2020.

\bibitem{Chen2020Improved}
Xinlei Chen, Haoqi Fan, Ross Girshick, and Kaiming He.
\newblock Improved baselines with momentum contrastive learning.
\newblock {\em arXiv}, 2020.

\bibitem{Dai2019Second}
Tao Dai, Jianrui Cai, Yongbing Zhang, Shu-Tao Xia, and Lei Zhang.
\newblock Second-order attention network for single image super-resolution.
\newblock In {\em CVPR}, 2019.

\bibitem{Doersch2015Unsupervised}
Carl Doersch, Abhinav Gupta, and Alexei~A Efros.
\newblock Unsupervised visual representation learning by context prediction.
\newblock In {\em ICCV}, pages 1422--1430, 2015.

\bibitem{Dong2014Learning}
Chao Dong, Chen~Change Loy, Kaiming He, and Xiaoou Tang.
\newblock Learning a deep convolutional network for image super-resolution.
\newblock In {\em ECCV}, pages 184--199, 2014.

\bibitem{Dosovitskiy2014Discriminative}
Alexey Dosovitskiy, Jost~Tobias Springenberg, Martin Riedmiller, and Thomas
  Brox.
\newblock Discriminative unsupervised feature learning with convolutional
  neural networks.
\newblock In {\em NeurIPS}, pages 766--774, 2014.

\bibitem{Gidaris2018Unsupervised}
Spyros Gidaris, Praveer Singh, and Nikos Komodakis.
\newblock Unsupervised representation learning by predicting image rotations.
\newblock {\em ICLR}, 2018.

\bibitem{Gu2019Blind}
Jinjin Gu, Hannan Lu, Wangmeng Zuo, and Chao Dong.
\newblock Blind super-resolution with iterative kernel correction.
\newblock In {\em CVPR}, 2019.

\bibitem{Hadsell2006Dimensionality}
Raia Hadsell, Sumit Chopra, and Yann LeCun.
\newblock Dimensionality reduction by learning an invariant mapping.
\newblock In {\em CVPR}, 2006.

\bibitem{Haris2018Deep}
Muhammad Haris, Gregory Shakhnarovich, and Norimichi Ukita.
\newblock Deep back-projection networks for super-resolution.
\newblock In {\em CVPR}, pages 1664--1673, 2018.

\bibitem{He2019Modulating}
Jingwen He, Chao Dong, and Yu Qiao.
\newblock Modulating image restoration with continual levels via adaptive
  feature modification layers.
\newblock In {\em CVPR}, pages 11056--11064, 2019.

\bibitem{He2020Interactive}
Jingwen He, Chao Dong, and Yu Qiao.
\newblock Interactive multi-dimension modulation with dynamic controllable
  residual learning for image restoration.
\newblock In {\em ECCV}, 2020.

\bibitem{He2020Momentum}
Kaiming He, Haoqi Fan, Yuxin Wu, Saining Xie, and Ross~B. Girshick.
\newblock Momentum contrast for unsupervised visual representation learning.
\newblock In {\em CVPR}, 2020.

\bibitem{Henaff2019Data}
Olivier~J H{\'e}naff, Aravind Srinivas, Jeffrey De~Fauw, Ali Razavi, Carl
  Doersch, SM Eslami, and Aaron van~den Oord.
\newblock Data-efficient image recognition with contrastive predictive coding.
\newblock In {\em CVPR}, 2019.

\bibitem{Huang2015Single}
Jia{-}Bin Huang, Abhishek Singh, and Narendra Ahuja.
\newblock Single image super-resolution from transformed self-exemplars.
\newblock In {\em CVPR}, pages 5197--5206, 2015.

\bibitem{Hussein2020Correction}
Shady~Abu Hussein, Tom Tirer, and Raja Giryes.
\newblock Correction filter for single image super-resolution: Robustifying
  off-the-shelf deep super-resolvers.
\newblock In {\em CVPR}, 2020.

\bibitem{Kim2016Accurate}
Jiwon Kim, Jung~Kwon Lee, and Kyoung~Mu Lee.
\newblock Accurate image super-resolution using very deep convolutional
  networks.
\newblock In {\em CVPR}, pages 1646--1654, 2016.

\bibitem{Kim2016Deeply}
Jiwon Kim, Jung~Kwon Lee, and Kyoung~Mu Lee.
\newblock Deeply-recursive convolutional network for image super-resolution.
\newblock In {\em CVPR}, pages 1637--1645, 2016.

\bibitem{Kingma2015Adam}
Diederik~P. Kingma and Jimmy Ba.
\newblock Adam: {A} method for stochastic optimization.
\newblock In {\em ICLR}, 2015.

\bibitem{Lim2017Enhanced}
Bee Lim, Sanghyun Son, Heewon Kim, Seungjun Nah, and Kyoung~Mu Lee.
\newblock Enhanced deep residual networks for single image super-resolution.
\newblock In {\em CVPR}, 2017.

\bibitem{Luo2020Unfolding}
Zhengxiong Luo, Yang Huang, Shang Li, Liang Wang, and Tieniu Tan.
\newblock Unfolding the alternating optimization for blind super resolution.
\newblock In {\em NeurIPS}, 2020.

\bibitem{Maaten2008Visualizing}
Laurens van~der Maaten and Geoffrey Hinton.
\newblock Visualizing data using t-{SNE}.
\newblock {\em Journal of Machine Learning Research}, 9(Nov):2579--2605, 2008.

\bibitem{Martin2001database}
David Martin, Charless Fowlkes, Doron Tal, Jitendra Malik, et~al.
\newblock A database of human segmented natural images and its application to
  evaluating segmentation algorithms and measuring ecological statistics.
\newblock In {\em ICCV}, 2001.

\bibitem{Michaeli2013Nonparametric}
Tomer Michaeli and Michal Irani.
\newblock Nonparametric blind super-resolution.
\newblock In {\em ICCV}, pages 945--952, 2013.

\bibitem{Noroozi2017Representation}
Mehdi Noroozi, Hamed Pirsiavash, and Paolo Favaro.
\newblock Representation learning by learning to count.
\newblock In {\em ICCV}, pages 5898--5906, 2017.

\bibitem{Oord2018Representation}
Aaron van~den Oord, Yazhe Li, and Oriol Vinyals.
\newblock Representation learning with contrastive predictive coding.
\newblock {\em arXiv}, 2018.

\bibitem{Park2020Contrastive}
Taesung Park, Alexei~A. Efros, Richard Zhang, and Jun-Yan Zhu.
\newblock Contrastive learning for unpaired image-to-image translation.
\newblock In {\em ECCV}, 2020.

\bibitem{Qiu2019Embedded}
Yajun Qiu, Ruxin Wang, Dapeng Tao, and Jun Cheng.
\newblock Embedded block residual network: A recursive restoration model for
  single-image super-resolution.
\newblock In {\em ICCV}, 2019.

\bibitem{Shocher2018Zero}
Assaf Shocher, Nadav Cohen, and Michal Irani.
\newblock ``{Z}ero-shot" super-resolution using deep internal learning.
\newblock In {\em CVPR}, 2018.

\bibitem{Soh2020Meta}
Jae~Woong Soh, Sunwoo Cho, and Nam~Ik Cho.
\newblock Meta-transfer learning for zero-shot super-resolution.
\newblock In {\em CVPR}, 2020.

\bibitem{Tian2019Contrastive}
Yonglong Tian, Dilip Krishnan, and Phillip Isola.
\newblock Contrastive multiview coding.
\newblock {\em arXiv}, 2019.

\bibitem{Timofte2017NTIRE}
Radu Timofte, Eirikur Agustsson, Luc~Van Gool, Ming{-}Hsuan Yang, Lei Zhang,
  and Bee~Lim \emph{et al.}
\newblock {NTIRE} 2017 challenge on single image super-resolution: Methods and
  results.
\newblock In {\em CVPRW}, pages 1110--1121, 2017.

\bibitem{Wu2018Unsupervised}
Zhirong Wu, Yuanjun Xiong, Stella~X Yu, and Dahua Lin.
\newblock Unsupervised feature learning via non-parametric instance
  discrimination.
\newblock In {\em CVPR}, pages 3733--3742, 2018.

\bibitem{Xu2020Unified}
Yu-Syuan Xu, Shou-Yao~Roy Tseng, Yu Tseng, Hsien-Kai Kuo, and Yi-Min Tsai.
\newblock Unified dynamic convolutional network for super-resolution with
  variational degradations.
\newblock In {\em CVPR}, pages 12496--12505, 2020.

\bibitem{Zeyde2010Single}
Roman Zeyde, Michael Elad, and Matan Protter.
\newblock On single image scale-up using sparse-representations.
\newblock In {\em International Conference on Curves and Surfaces}, volume
  6920, pages 711--730, 2010.

\bibitem{Zhang2020Deep}
Kai Zhang, Luc~Van Gool, and Radu Timofte.
\newblock Deep unfolding network for image super-resolution.
\newblock In {\em CVPR}, 2020.

\bibitem{Zhang2017Gaussian}
Kai Zhang, Wangmeng Zuo, Yunjin Chen, Deyu Meng, and Lei Zhang.
\newblock Beyond a gaussian denoiser: Residual learning of deep {CNN} for image
  denoising.
\newblock {\em {IEEE} Trans. Image Process.}, 26(7):3142--3155, jul 2017.

\bibitem{Zhang2017Learning}
Kai Zhang, Wangmeng Zuo, and Lei Zhang.
\newblock Learning a single convolutional super-resolution network for multiple
  degradations.
\newblock In {\em CVPR}, 2017.

\bibitem{Zhang2016Colorful}
Richard Zhang, Phillip Isola, and Alexei~A Efros.
\newblock Colorful image colorization.
\newblock In {\em ECCV}, pages 649--666, 2016.

\bibitem{Zhang2018Image}
Yulun Zhang, Kunpeng Li, Kai Li, Lichen Wang, Bineng Zhong, and Yun Fu.
\newblock Image super-resolution using very deep residual channel attention
  networks.
\newblock In {\em ECCV}, pages 1646--1654, 2018.

\bibitem{Zhang2018Residual}
Yulun Zhang, Yapeng Tian, Yu Kong, Bineng Zhong, and Yun Fu.
\newblock Residual dense network for image super-resolution.
\newblock In {\em CVPR}, pages 2472--2481, 2018.

\end{thebibliography}


\begin{thebibliography}{1}\itemsep=-1pt

\bibitem{Gu2019Blind}
Jinjin Gu, Hannan Lu, Wangmeng Zuo, and Chao Dong.
\newblock Blind super-resolution with iterative kernel correction.
\newblock In {\em CVPR}, 2019.

\bibitem{Hussein2020Correction}
Shady~Abu Hussein, Tom Tirer, and Raja Giryes.
\newblock Correction filter for single image super-resolution: Robustifying
  off-the-shelf deep super-resolvers.
\newblock In {\em CVPR}, 2020.

\bibitem{Luo2020Unfolding}
Zhengxiong Luo, Yang Huang, Shang Li, Liang Wang, and Tieniu Tan.
\newblock Unfolding the alternating optimization for blind super resolution.
\newblock In {\em NeurIPS}, 2020.

\bibitem{Soh2020Meta}
Jae~Woong Soh, Sunwoo Cho, and Nam~Ik Cho.
\newblock Meta-transfer learning for zero-shot super-resolution.
\newblock In {\em CVPR}, 2020.

\bibitem{Zhang2020Deep}
Kai Zhang, Luc~Van Gool, and Radu Timofte.
\newblock Deep unfolding network for image super-resolution.
\newblock In {\em CVPR}, 2020.

\bibitem{Zhang2018Image}
Yulun Zhang, Kunpeng Li, Kai Li, Lichen Wang, Bineng Zhong, and Yun Fu.
\newblock Image super-resolution using very deep residual channel attention
  networks.
\newblock In {\em ECCV}, pages 1646--1654, 2018.

\end{thebibliography}
	}
	
\end{document}